\documentclass[11pt]{article}

\usepackage{acl}

\usepackage{times}
\usepackage{latexsym}

\usepackage[T1]{fontenc}

\usepackage[utf8]{inputenc}

\usepackage{microtype}

\usepackage{inconsolata}

\usepackage{graphicx}
\usepackage{booktabs}
\usepackage{makecell}
\usepackage{multirow} 
\usepackage{tabularx}
\usepackage{array}
\usepackage{amsmath}
\usepackage{amssymb}
\title{Preserve to Expand: Source-Aware Adaptation and Model Merging for Multilingual LLM-Based ASR}

\author{
Yuchen Zhang\textsuperscript{1,2}  \and Haralambos Mouratidis\textsuperscript{1,2} \and Ravi Shekhar\textsuperscript{1,2}\\
\textsuperscript{1}Institute for Analytics and Data Science, University of Essex \\
\textsuperscript{2}School of Computer Science and Electronic Engineering, University of Essex \\  
\texttt{\{yuchen.zhang, h.mouratidis, r.shekhar\}@essex.ac.uk}
}

\begin{document}
\maketitle
\begin{abstract}
Expanding multilingual ASR systems to new languages is challenging when adaptation must preserve previously supported languages without replaying source-language training data. We study this problem in connector-based Speech LLMs, where a trainable connector maps speech representations into a frozen large language model. We evaluate target-update construction and merging strategies for adapting the connector while preserving source-language performance. In single-target expansion, target-only fine-tuning and standard TIES merging learn new languages but cause severe source-language forgetting, while SSU-Wanda+TIES gives the best source-preserving trade-off. In multi-target expansion, update merging provides stable source retention, whereas recursive expansion can improve target adaptation under favorable orders but is highly order-sensitive. These results show that multilingual ASR expansion requires balancing target acquisition, source retention, and interactions among target-language updates.
\end{abstract}

\section{Introduction}

Recent speech foundation models have substantially improved multilingual automatic speech recognition (ASR) through larger training corpora, increased model capacity, and improved training objectives \cite{babu2022xlsr,radford2022whisper,barrault2023seamlessm4t}. However, extending an existing multilingual ASR model to new languages remains challenging. Expansion often needs to be performed with limited data and computation while preserving performance on previously supported languages, creating a trade-off between target-language adaptation and source-language retention.

A common expansion strategy is to fine-tune the existing model on the new language, but this can cause catastrophic forgetting. Similar degradation has been observed in connector-based Speech LLMs for ASR, where multilingual projectors are easy to adapt but may substantially harm original-language performance \cite{concina2026meusli,concina2026scalable}. Replay-based continual learning can mitigate forgetting by interleaving previously supported languages during adaptation, but it requires storing and reusing replay data. When such data is unavailable or impractical to reuse, expansion must rely on replay-free target-language adaptation without gradient-based source adaptation.

Model merging offers an alternative by combining learned parameter updates without additional training~\cite{yang2026model}. Methods such as weight averaging and task arithmetic integrate capabilities from different training runs or tasks in weight space~\cite{wortsman2022model,ilharco2022editing}, but independently learned updates may interfere. TIES-Merging mitigates this interference by removing small parameter changes and resolving sign conflicts between task vectors~\cite{yadav2023ties}. Language-specific merging has also been explored to reduce training and maintenance costs when adding or updating languages in multilingual models~\cite{dmonte2026improving}. However, how target-language updates behave in connector-based multilingual ASR expansion remains unclear, particularly without source-language replay during adaptation.

In this work, we study source-aware multilingual ASR expansion for connector-based Speech LLMs~\cite{ma2025speech, ma2024simpleasr}. Starting from a multilingual base connector~\cite{zhang-etal-2026-language}, our goal is to adapt the connector to new target languages while preserving performance on the source languages, without replaying source-language training data or performing gradient-based source adaptation during expansion. We focus on two questions: \textit{RQ1: How can adaptation and merging balance target acquisition and source retention?} and \textit{RQ2: How effectively can independent target updates be merged to support multiple new languages?}

To answer these questions, we conduct controlled experiments with three multilingual base connectors and six target languages. We compare direct target fine-tuning, source-aware parameter-selective adaptation, and TIES-based target-update merging under the same target-language adaptation budget.  Our experimental results suggest that source-aware merging substantially improves the adaptation--retention trade-off in single-target expansion, while multi-target expansion requires managing interactions among target-language updates. Joint update merging provides stable source retention, whereas recursive adaptation can achieve stronger target performance under favorable target orders.

Our main contributions are as follows. First, we introduce a source-aware multilingual ASR expansion setting for connector-based Speech LLMs, where new target languages are added while preserving existing source-language performance without source-language replay or gradient-based source adaptation. Second, we systematically compare target-update construction strategies, including direct fine-tuning, SSU-based parameter-selective adaptation, SSU+FT, and TIES-based target-update merging, to characterize the trade-off between target adaptation and source retention. Third, we extend the study from single-target to multi-target expansion and analyze how target-update interactions and recursive target order shape multilingual expansion.

\section{Related Work}

\subsection{Multilingual ASR Expansion}

Large-scale speech foundation models have advanced multilingual ASR through cross-lingual self-supervised learning, weakly supervised recognition, and unified multilingual speech modeling across diverse languages and tasks \cite{babu2022xlsr,radford2022whisper,barrault2023seamlessm4t}. Recent LLM-based ASR systems further connect pretrained speech encoders with large language models through lightweight connectors, showing that frozen speech and language backbones can support multilingual recognition and benefit from auxiliary multimodal context \cite{fathullah2023prompting,ma2024simpleasr,yang2024malaasr}. These developments motivate connector-based Speech LLMs as a parameter-efficient ASR framework, but leave open how to expand them to new languages while preserving existing multilingual capabilities.

Multilingual ASR systems often need to be expanded after the initial model has been trained, making new-language acquisition and source-language retention a central challenge. Prior work on multilingual ASR expansion has largely framed new-language acquisition and forgetting as a continual-learning problem, covering replay-based retention, parameter-regularized adaptation, and decoder-specific optimization strategies \cite{della2024cl,kwok24_interspeech,concina2026scalable}. In contrast, our work does not rely on source-language replay or gradient-based source adaptation during expansion. We instead study how target-language updates can be constructed and combined under lightweight source-language calibration, comparing parameter-selective adaptation and target-update merging in both single-target and multi-target connector expansion.

\subsection{Model Merging for Adaptation}

Model merging provides an alternative to continued fine-tuning by combining parameters or parameter updates from separately trained models. Model soups show that averaging the weights of multiple fine-tuned models can improve performance and robustness without increasing inference cost, particularly when the models are fine-tuned from the same initialization \cite{wortsman2022model}. Task arithmetic represents fine-tuning effects as task vectors and shows that these vectors can be added, subtracted, or composed to alter model behavior in parameter space \cite{ilharco2022editing}. These approaches suggest that independently learned capabilities can sometimes be integrated without additional training.

However, merging fine-tuned models separately can introduce interference between parameter updates. TIES-Merging addresses this issue by trimming small parameter changes, resolving sign conflicts among task vectors, and merging only sign-aligned updates \cite{yadav2023ties}. In multilingual settings, \citet{dmonte2026improving} study language-specific model merging as a way to reduce training and maintenance costs when individual languages are added or updated in multilingual task-specific LLMs. In the speech domain, model merging is explored for adding new tasks~\cite{zhao2026low,lin2025speech}, domains~\cite{carvalho2026exploring}, and fairness~\cite{lin2026pseudo2real,bravo2026intercorrect}.  Nevertheless, existing model merging work has not fully examined connector-based multilingual ASR expansion under a source-free setting, where target-language updates must be integrated while preserving source-language performance without access to source-language data during expansion.

\subsection{Parameter-Selective Adaptation}
Parameter-selective adaptation aims to reduce forgetting by restricting which parameters are updated during task or language adaptation~\cite{liu2021autofreeze,zhang-etal-2024-balancing,pan2024lisa}. Instead of updating the full model, these methods identify parameters or structural units that are important for previously learned capabilities and protect them during subsequent fine-tuning. This idea has been explored in continual learning through masking, pruning, and parameter-isolation strategies, where different subsets of model parameters are assigned to different tasks or protected from future updates \cite{mallya2018packnet, mallya2018piggyback}.

Source-Shielded Updates (SSU) apply this idea to target-language adaptation of LLMs by estimating source-language parameter importance and freezing source-important units during target-language fine-tuning \cite{yamaguchi2026mitigating}. This is particularly relevant to multilingual ASR expansion, where adapting to a new language can interfere with previously supported source languages. In our work, we adopt SSU-style source-aware masking for connector adaptation, using a lightweight source-language calibration set only to construct masks. 

\section{Methodology}
\label{sec:method}

\begin{figure*}[!t]
\centering

\includegraphics[width=\textwidth,height=0.31\textheight]{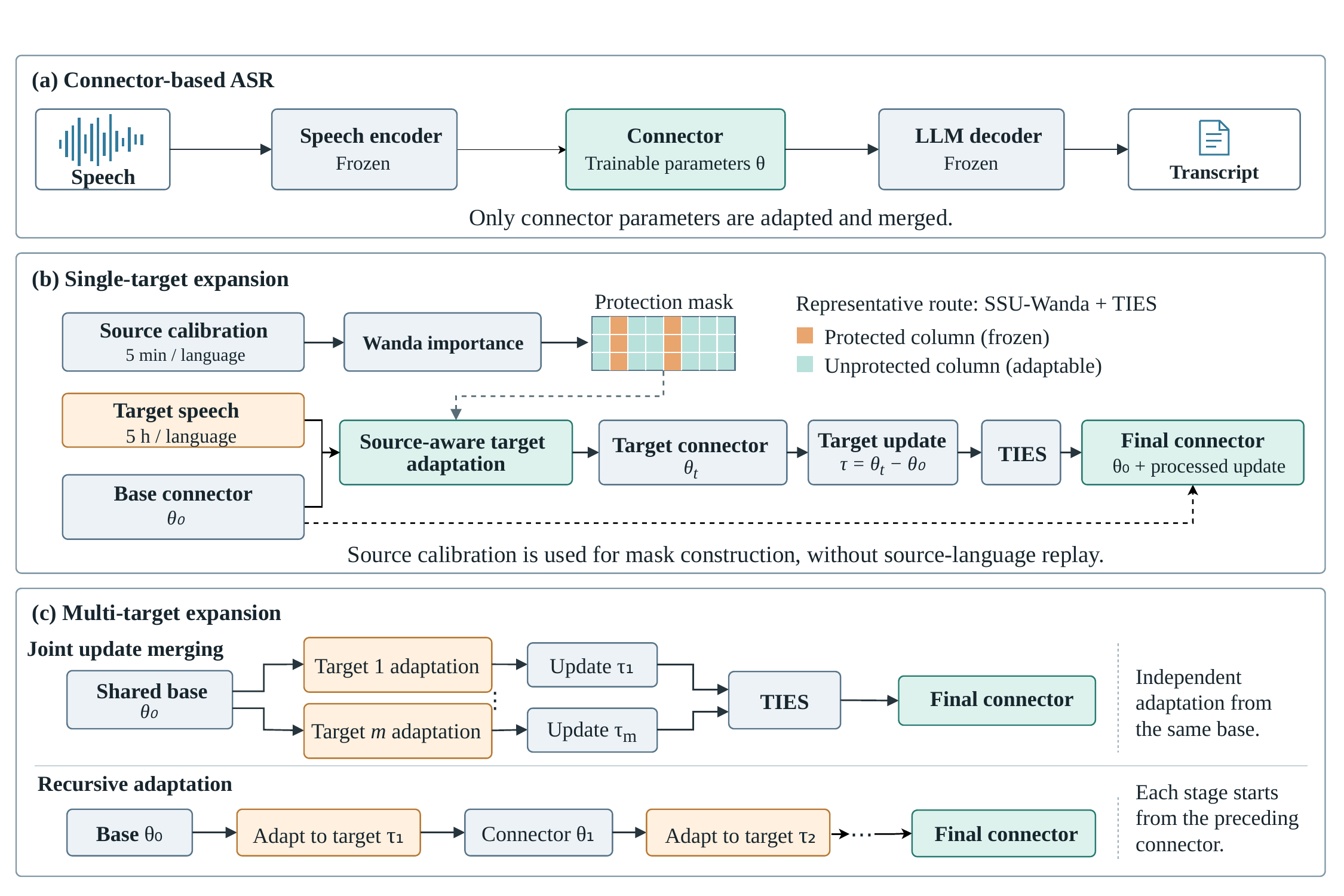}

\caption{Overview of source-aware multilingual ASR expansion. }
\label{fig:framework}
\end{figure*}

Figure~\ref{fig:framework} provides an overview of the connector-based ASR architecture and the single-target and multi-target expansion strategies studied in this work.

\subsection{Connector-Based ASR Backbone}
\label{sec:backbone}

We study multilingual expansion in a connector-based ASR architecture~\cite{ma2025speech, ma2024simpleasr, zhang-etal-2026-language}, where only a lightweight connector is adapted while both the speech encoder and the language model remain frozen. Given an input speech signal $S$, a frozen speech encoder $\mathcal{E}$ produces acoustic representations:
\begin{equation}
\mathbf{H} = \mathcal{E}(S) \in \mathbb{R}^{B \times T \times E},
\end{equation}
where $B$ is the batch size, $T$ is the number of acoustic frames, and $E$ is the encoder hidden dimension. To reduce the frame rate before passing the representations to the LLM, we apply a stack-downsampling operator $\mathcal{D}_K(\cdot)$ that concatenates every $K$ consecutive frames:
\begin{equation}
\mathbf{H}^{(K)} = \mathcal{D}_K(\mathbf{H}) \in \mathbb{R}^{B \times (T/K) \times (EK)}.
\end{equation}

A trainable connector $\mathcal{C}_{\theta}$ maps the downsampled acoustic sequence into the LLM embedding space:
\begin{equation}
\mathbf{Z} = \mathcal{C}_{\theta}(\mathbf{H}^{(K)}),
\end{equation}
where $\theta$ denotes the connector parameters. The frozen decoder-only LLM $\mathcal{D}$ then generates the transcription autoregressively:
\begin{equation}
p(\mathbf{y}\mid S) = \prod_{t=1}^{|\mathbf{y}|} p(y_t \mid y_{<t}, \mathbf{Z}).
\end{equation}
All adaptation and merging operations in this work are applied only to the connector parameters, while the speech encoder and LLM decoder remain frozen.

\subsection{Target Adaptation with Source Calibration}
\label{sec:adaptation}

We begin with a multilingual base connector trained on a set of source languages. Let $\theta_{\mathrm{base}}$ denote the resulting base connector. Given a new target language, our goal is to adapt the connector to the target language while retaining performance on the original source languages.

We first construct a target-adapted connector by fine-tuning the base connector on target-language data:
\begin{equation}
\theta_{\mathrm{tgt}} = \mathrm{FT}(\theta_{\mathrm{base}}, \mathcal{D}_{\mathrm{tgt}}),
\end{equation}
where $\mathcal{D}_{\mathrm{tgt}}$ denotes the target-language adaptation data. Direct fine-tuning uses $\theta_{\mathrm{tgt}}$ itself as the expanded model. In merging-based settings, we view target adaptation as an update from the base connector:
\begin{equation}
\tau_{\mathrm{tgt}} = \theta_{\mathrm{tgt}} - \theta_{\mathrm{base}}.
\end{equation}

We also use SSU as a parameter-selective target adaptation strategy \cite{yamaguchi2026mitigating}. SSU uses a small source-language calibration set to estimate which columns of each weight matrix are important for source-language behavior, and then freezes those columns during target-language adaptation. In our work, the calibration set is drawn from the source languages of the base connector. 

Let $\rho$ denote the column-freezing ratio. For each connector weight matrix $\mathbf{W}\in\mathbb{R}^{d_{\mathrm{out}}\times d_{\mathrm{in}}}$, SSU first assigns an element-wise importance score $s_{ij}$ to each weight $W_{ij}$. Following SSU, we use the Wanda criterion to compute these scores from source-language calibration data. Specifically,
\begin{equation}
s_{ij}^{\mathrm{Wanda}}
=
|W_{ij}| \cdot \|\mathbf{x}_j\|_2,
\end{equation}
where $\mathbf{x}_j$ is the activation of the $j$-th input feature estimated from source-language calibration data.

We also evaluate two alternative masking criteria. In \textsc{SSU-Mag}, the element-wise score is based only on weight magnitude:
\begin{equation}
s_{ij}^{\mathrm{Mag}} = |W_{ij}|.
\end{equation}
In \textsc{SSU-Rand}, columns are selected uniformly at random, without using an importance score.

For score-based variants, SSU aggregates element-wise scores into column scores:
\begin{equation}
S_j = \sum_{i=1}^{d_{\mathrm{out}}} s_{ij}.
\end{equation}
The protected column set is selected by taking the top-$\rho$ proportion of columns:
\begin{equation}
\mathcal{P}_{\rho}
=
\operatorname{Top}_{\rho}
\left(
\{S_j\}_{j=1}^{d_{\mathrm{in}}}
\right).
\end{equation}
For \textsc{SSU-Rand}, $\mathcal{P}_{\rho}$ is instead sampled uniformly from the columns of $\mathbf{W}$ with the same cardinality.

The binary mask for $\mathbf{W}$ is defined as
\begin{equation}
B_{ij}
=
\begin{cases}
0, & \text{if } j \in \mathcal{P}_{\rho}, \\
1, & \text{if } j \notin \mathcal{P}_{\rho}.
\end{cases}
\end{equation}
Thus, protected columns are frozen during target-language adaptation:
\begin{equation}
W_{ij}
\leftarrow
W_{ij}
-
\eta B_{ij}\nabla_{W_{ij}}\mathcal{L}_{\mathrm{tgt}},
\end{equation}
where $\eta$ is the learning rate and $\mathcal{L}_{\mathrm{tgt}}$ is the target-language training loss. For one-dimensional parameters such as biases, each parameter is treated as its own column.

The resulting SSU-adapted connector is denoted as
\begin{equation}
\theta_{\mathrm{tgt}}^{\mathrm{SSU}}
=
\mathrm{FT}_{\mathbf{B}}(\theta_{\mathrm{base}}, \mathcal{D}_{\mathrm{tgt}}),
\end{equation}
with update
\begin{equation}
\tau_{\mathrm{tgt}}^{\mathrm{SSU}}
=
\theta_{\mathrm{tgt}}^{\mathrm{SSU}} - \theta_{\mathrm{base}}.
\end{equation}

Finally, we include an \textsc{SSU+FT} variant. In this setting, SSU is first used to obtain $\theta_{\mathrm{tgt}}^{\mathrm{SSU}}$, which is then further adapted with standard unmasked target-language fine-tuning:
\begin{equation}
\theta_{\mathrm{tgt}}^{\mathrm{SSU+FT}}
=
\mathrm{FT}(\theta_{\mathrm{tgt}}^{\mathrm{SSU}}, \mathcal{D}_{\mathrm{tgt}}).
\end{equation}
This variant tests whether source-calibrated masked adaptation provides a better initialization for subsequent target-language optimization.

\subsection{Target-Update Model Merging}
\label{sec:merging}


For single-target expansion, the target update is
\begin{equation}
\tau_{\mathrm{tgt}} = \theta_{\mathrm{tgt}} - \theta_{\mathrm{base}},
\end{equation}
where $\theta_{\mathrm{tgt}}$ may be obtained from standard fine-tuning, SSU-based adaptation, or SSU+FT. We instantiate the merging operator using TIES-Merging \cite{yadav2023ties}. Given a set of target updates
\begin{equation}
\mathcal{T} = \{\tau^{(1)}, \ldots, \tau^{(n)}\},
\end{equation}
where each $\tau^{(j)}\in\mathbb{R}^{d}$ is defined relative to $\theta_{\mathrm{base}}$, TIES produces a merged update $\tau_m$ through trim, elect sign, and disjoint merge.

First, TIES trims each update by retaining only the largest $k$\% of entries by absolute value:
\begin{equation}
\hat{\tau}^{(j)}_i
=
\begin{cases}
\tau^{(j)}_i, & \text{if } |\tau^{(j)}_i| \ge q_k(|\tau^{(j)}|), \\
0, & \text{otherwise}.
\end{cases}
\end{equation}
Here, $q_k(|\tau^{(j)}|)$ is the threshold corresponding to the top-$k$\% absolute values in $\tau^{(j)}$. This removes small parameter changes before aggregation.

Second, TIES elects a sign for each parameter dimension based on the aggregate signed mass:
\begin{equation}
\gamma_i
=
\operatorname{sgn}
\left(
\sum_{j=1}^{n} \hat{\tau}^{(j)}_i
\right).
\end{equation}
When the signed mass is zero, we set $\gamma_i=0$ and the merged update for that dimension to zero.

Finally, TIES performs a disjoint merge by averaging only updates whose signs agree with the elected sign. Let
\begin{equation}
\mathcal{A}_i
=
\left\{
j \; \middle| \;
\operatorname{sgn}(\hat{\tau}^{(j)}_i) = \gamma_i
\;\text{and}\;
\hat{\tau}^{(j)}_i \neq 0
\right\}.
\end{equation}
The merged update is
\begin{equation}
(\tau_m)_i
=
\begin{cases}
\dfrac{1}{|\mathcal{A}_i|}
\sum_{j \in \mathcal{A}_i}
\hat{\tau}^{(j)}_i,
& \text{if } |\mathcal{A}_i| > 0, \\
0,
& \text{if } |\mathcal{A}_i| = 0.
\end{cases}
\end{equation}
The final merged connector is obtained by scaling the merged update and adding it to the base connector:
\begin{equation}
\theta_m = \theta_{\mathrm{base}} + \lambda \tau_m,
\end{equation}
where $\lambda$ is a scaling coefficient.

For conventional single-target merging, $\mathcal{T}=\{\tau_{\mathrm{tgt}}\}$. In this case, there are no cross-update sign conflicts, and TIES reduces to trimming the target update before scaling:
\begin{equation}
\theta_m
=
\theta_{\mathrm{base}}
+
\lambda \operatorname{TopK}(\tau_{\mathrm{tgt}}; k).
\end{equation}

For multi-target expansion, let $\tau_{\mathrm{tgt}}^{(1)}, \ldots, \tau_{\mathrm{tgt}}^{(m)}$ denote target-language updates independently obtained from the same base connector. We merge these target updatesly:
\begin{equation}
\tau_m
=
\mathcal{M}_{\mathrm{TIES}}
\left(
\{\tau_{\mathrm{tgt}}^{(1)}, \ldots, \tau_{\mathrm{tgt}}^{(m)}\}
\right),
\end{equation}
and obtain
\begin{equation}
\theta_{\mathrm{multi}}
=
\theta_{\mathrm{base}}
+
\lambda \tau_m.
\end{equation}
In this setting, TIES can resolve sign conflicts among multiple target-language updates before aggregation. This allows us to test whether independently learned target-language updates can be combined without recursive fine-tuning on multiple target languages.



\section{Experiments}
\label{sec:exp}


\subsection{Experimental Setting}
\label{sec:exp_setting}

We use Common Voice~\cite{ardila2020common} for base training and target adaptation. Following \citet{zhang2026language}, we train three multilingual base connectors covering Germanic, Romance, and Slavic languages and evaluate expansion to six target languages. Table~\ref{tab:lang_setup} lists the languages. Base training and target adaptation each use 5 hours per language. SSU-Wanda additionally uses 5 minutes per source language from the validation split for activation collection and mask construction, without source-language replay or gradient-based source adaptation.

\begin{table}[htbp]
\centering
\small
\resizebox{\linewidth}{!}{%
\begin{tabular}{ll}
\toprule
\textbf{Group} & \textbf{Languages} \\
\midrule
Germanic (GEM) &
English (EN), German (DE), Dutch (NL), Danish (DA), Swedish (SV) \\
Romance (ROA) &
French (FR), Spanish (ES), Portuguese (PT), Galician (GL), Catalan (CA) \\
Slavic (SLA) &
Russian (RU), Ukrainian (UK), Belarusian (BE), Bulgarian (BG), Macedonian (MK) \\
\midrule
Targets &
Norwegian (NO), Italian (IT), Serbian (SR), Polish (PL), Slovenian (SL), Kyrgyz (KY) \\
\bottomrule
\end{tabular}%
}
\caption{Source-language groups and target languages.}
\label{tab:lang_setup}
\end{table}

\begin{table}[htbp]
\centering
\small
\begin{tabular}{lll}
\toprule
\textbf{Type} & \textbf{Base} & \textbf{Target} \\
\midrule
\multirow{3}{*}{G+/S+}
& Germanic & Norwegian (NO) \\
& Romance & Italian (IT) \\
& Slavic & Serbian (SR) \\
\midrule
\multirow{2}{*}{G+/S-}
& Slavic & Polish (PL) \\
& Slavic & Slovenian (SL) \\
\midrule
\multirow{3}{*}{G-/S+}
& Germanic & Italian (IT) \\
& Romance & Norwegian (NO) \\
& Slavic & Kyrgyz (KY) \\
\midrule
\multirow{3}{*}{G-/S-}
& Germanic & Serbian (SR) \\
& Romance & Serbian (SR) \\
& Slavic & Italian (IT) \\
\bottomrule
\end{tabular}
\caption{Single-target base--target configurations.G+/G- indicate language-group
match/mismatch, and S+/S- indicate script match/mismatch.}
\label{tab:base_target_settings}
\end{table}

Single-target experiments add each target independently under language-group and script match/mismatch conditions, as listed in Table~\ref{tab:base_target_settings}. Multi-target experiments compare joint update merging with recursive adaptation, using Wanda-based masking for SSU. We report source- and target-language WER to assess retention and adaptation, respectively.

For all experiments, only connector parameters are adapted, while the speech encoder and LLM remain frozen.We use AdamW with a learning rate of $1\times10^{-4}$, weight decay of $1\times10^{-6}$, batch size 4, and downsampling factor $K=4$. Training runs for up to 5 epochs with early stopping on validation loss. We use the final checkpoint from each run and decode with beam size 4. All experiments share these training settings and are conducted on two NVIDIA A6000 GPUs. Further details of the experimental setup are provided in Appendix~\ref{app:experimental_details}.

\subsection{Single-Target Expansion}
\label{sec:single_target}

\subsubsection{Target Adaptation and Source Retention}

\begin{table}[!htbp]
\centering
\small
\setlength{\tabcolsep}{4pt}
\resizebox{\linewidth}{!}{%
\begin{tabular}{llrrrr}
\toprule
\textbf{Base} & \textbf{Method} & \textbf{Source} & \textbf{Target} & \textbf{All} & $\boldsymbol{\Delta}$\textbf{Source} \\
\midrule
\multirow{7}{*}{GEM}
& Base & 22.57 & 102.70 & 35.93 & +0.00 \\
& Target FT & 69.73 & \textbf{27.86} & 62.75 & +47.15 \\
& SSU-Wanda+FT & 67.46 & 28.47 & 60.96 & +44.89 \\
& TIES & 69.08 & 27.71 & 62.18 & +46.50 \\
& SSU-Rand+TIES & 29.42 & 30.30 & 29.56 & +6.84 \\
& SSU-Mag+TIES & 40.83 & 29.52 & 38.95 & +18.26 \\
& SSU-Wanda+TIES & \textbf{22.81} & 35.82 & \textbf{24.98} & \textbf{+0.24} \\
\midrule
\multirow{7}{*}{ROA}
& Base & 18.95 & 111.16 & 34.32 & +0.00 \\
& Target FT & 74.50 & 28.06 & 66.76 & +55.54 \\
& SSU-Wanda+FT & 55.15 & 29.45 & 50.87 & +36.20 \\
& TIES & 73.59 & \textbf{27.77} & 65.95 & +54.63 \\
& SSU-Rand+TIES & 33.79 & 29.50 & 33.08 & +14.84 \\
& SSU-Mag+TIES & 41.21 & 29.31 & 39.22 & +22.25 \\
& SSU-Wanda+TIES & \textbf{19.64} & 37.36 & \textbf{22.59} & \textbf{+0.69} \\
\midrule
\multirow{7}{*}{SLA}
& Base & 36.93 & 110.15 & 49.13 & +0.00 \\
& Target FT & 91.41 & 36.37 & 82.24 & +54.49 \\
& SSU-Wanda+FT & 85.61 & \textbf{36.15} & 77.36 & +48.68 \\
& TIES & 90.88 & 37.03 & 81.90 & +53.95 \\
& SSU-Rand+TIES & 59.46 & 41.69 & 56.50 & +22.54 \\
& SSU-Mag+TIES & 72.06 & 40.55 & 66.81 & +35.13 \\
& SSU-Wanda+TIES & \textbf{43.60} & 51.69 & \textbf{44.95} & \textbf{+6.68} \\
\bottomrule
\end{tabular}}
\caption{Macro-averaged WER (\%) $\downarrow$ for single-target expansion by base.}
\label{tab:single_summary_main}
\end{table}

Table~\ref{tab:single_summary_main} reports macro-averaged WER for each base group. The results show a clear adaptation--retention trade-off. Target-only fine-tuning learns the new language effectively, but severely damages the original source languages. For example, on the Germanic base, Target FT reduces target WER from 102.70\% to 27.86\%, but increases source WER from 22.57\% to 69.73\%. The same pattern appears for Romance and Slavic, where source WER increases by +55.54\% and +54.49\%, respectively.

Standard TIES merging does not solve this problem in the single-target setting. Although it preserves the target-language gains of fine-tuning, it retains most of the source-language degradation, behaving similarly to Target FT across base groups. SSU-Wanda+FT is also insufficient on its own, even with source-important parameters frozen during target adaptation, the resulting connector still suffers substantial source degradation. These results suggest that neither conventional merging nor protected fine-tuning alone is enough to recover a balanced source--target solution.

In contrast, SSU-Wanda+TIES gives the strongest overall trade-off. It keeps source WER close to the base connector while still improving target performance substantially. For Germanic and Romance, source WER increases by only +0.24\% and +0.69\%, while target WER drops from 102.70\% to 35.82\% and from 111.16\% to 37.36\%, respectively. For Slavic, the harder expansion setting, SSU-Wanda+TIES reduces source degradation from +53.95\% under TIES to +6.68\%, while improving target WER from 110.15\% to 51.69\%. These results show that strong single-target expansion requires combining source-aware adaptation with update merging.

\subsubsection{Effect of Source-Aware Protection}

We next compare the three SSU masking criteria used with TIES: random masking, magnitude-based masking, and Wanda-based masking.
Table~\ref{tab:single_summary_main} shows that all SSU+TIES variants reduce source forgetting compared with Target FT and standard TIES, but they differ in how they balance target adaptation and source retention. SSU-Rand+TIES often preserves stronger target-language gains than SSU-Wanda+TIES, but leaves more source degradation. For example, on the Slavic base, SSU-Rand+TIES reaches 41.69\% target WER, but increases source WER by +22.54\%. SSU-Wanda+TIES gives a higher target WER of 51.69\%, but reduces source degradation to +6.68\%. This suggests that random protection can limit update damage, but does not reliably select the directions most important for preserving source-language behavior.

SSU-Mag+TIES provides a more structured mask than random selection, but it is generally worse than SSU-Wanda+TIES in source retention and overall WER. Across all three base groups, SSU-Mag+TIES causes larger source degradation than SSU-Wanda+TIES, indicating that weight magnitude alone is not sufficient to identify source-important connector directions. By incorporating source-language activations, SSU-Wanda produces a more conservative mask that better preserves the original multilingual capability.

Overall, SSU-Wanda+TIES gives the most reliable source-preserving trade-off. It achieves the lowest source WER and lowest overall WER for all three base groups, although it sometimes gives up part of the target-language gain achieved by SSU-Rand or SSU-Mag. We therefore use SSU-Wanda as the default SSU criterion in the multi-target experiments.

\subsubsection{Base--Target Compatibility}

\begin{table}[!htbp]
\centering
\small
\setlength{\tabcolsep}{3pt}
\resizebox{\linewidth}{!}{%
\begin{tabular}{lllrrrr}
\toprule
\textbf{Target} & \textbf{Type} & \textbf{Method} & \textbf{Source} & \textbf{Target} & \textbf{All} & $\boldsymbol{\Delta}$\textbf{Source} \\
\midrule
\multirow{7}{*}{IT}
& \multirow{7}{*}{G-/S-}
& Base & 36.93 & 116.08 & 50.12 & +0.00 \\
& & Target FT & 129.79 & \textbf{20.11} & 111.51 & +92.86 \\
& & SSU-Wanda+FT & 115.24 & 23.36 & 99.93 & +78.31 \\
& & TIES & 128.32 & 21.07 & 110.45 & +91.40 \\
& & SSU-Rand+TIES & 52.82 & 39.52 & 50.60 & +15.89 \\
& & SSU-Mag+TIES & 82.62 & 34.28 & 74.56 & +45.70 \\
& & SSU-Wanda+TIES & \textbf{34.35} & 70.57 & \textbf{40.39} & \textbf{-2.57} \\
\midrule
\multirow{7}{*}{SR}
& \multirow{7}{*}{G+/S+}
& Base & 36.93 & 66.58 & 41.87 & +0.00 \\
& & Target FT & 57.12 & \textbf{24.51} & 51.69 & +20.19 \\
& & SSU-Wanda+FT & 55.56 & 24.62 & 50.40 & +18.63 \\
& & TIES & 56.46 & 24.62 & 51.15 & +19.53 \\
& & SSU-Rand+TIES & 44.26 & 26.03 & 41.22 & +7.33 \\
& & SSU-Mag+TIES & 46.12 & 25.91 & 42.75 & +9.19 \\
& & SSU-Wanda+TIES & \textbf{41.10} & 27.63 & \textbf{38.85} & \textbf{+4.17} \\
\midrule
\multirow{7}{*}{PL}
& \multirow{7}{*}{G+/S-}
& Base & 36.93 & 107.36 & 48.67 & +0.00 \\
& & Target FT & 107.05 & 26.39 & 93.61 & +70.13 \\
& & SSU-Wanda+FT & 104.80 & \textbf{26.17} & 91.70 & +67.87 \\
& & TIES & 107.36 & 26.28 & 93.85 & +70.43 \\
& & SSU-Rand+TIES & 76.95 & 28.09 & 68.81 & +40.02 \\
& & SSU-Mag+TIES & 95.16 & 27.42 & 83.87 & +58.24 \\
& & SSU-Wanda+TIES & \textbf{45.33} & 35.45 & \textbf{43.68} & \textbf{+8.41} \\
\midrule
\multirow{7}{*}{SL}
& \multirow{7}{*}{G+/S-}
& Base & 36.93 & 110.70 & 49.22 & +0.00 \\
& & Target FT & 105.60 & \textbf{38.47} & 94.41 & +68.67 \\
& & SSU-Wanda+FT & 99.34 & 38.84 & 89.26 & +62.41 \\
& & TIES & 105.37 & 40.07 & 94.49 & +68.44 \\
& & SSU-Rand+TIES & 81.99 & 39.99 & 74.99 & +45.07 \\
& & SSU-Mag+TIES & 92.93 & 40.03 & 84.11 & +56.00 \\
& & SSU-Wanda+TIES & \textbf{58.04} & 43.72 & \textbf{55.65} & \textbf{+21.11} \\
\midrule
\multirow{7}{*}{KY}
& \multirow{7}{*}{G-/S+}
& Base & 36.93 & 150.01 & 55.78 & +0.00 \\
& & Target FT & 57.51 & 72.39 & 59.99 & +20.58 \\
& & SSU-Wanda+FT & 53.09 & \textbf{67.77} & 55.54 & +16.16 \\
& & TIES & 56.89 & 73.09 & 59.59 & +19.96 \\
& & SSU-Rand+TIES & 41.30 & 74.84 & 46.89 & +4.37 \\
& & SSU-Mag+TIES & 43.45 & 75.09 & 48.72 & +6.52 \\
& & SSU-Wanda+TIES & \textbf{39.19} & 81.10 & \textbf{46.18} & \textbf{+2.26} \\
\bottomrule
\end{tabular}}
\caption{Detailed single-target expansion  WER (\%)$\downarrow$ for the Slavic base.}
\label{tab:single_compatibility_main}
\end{table}

Finally, we examine whether expansion difficulty depends on the relationship between the base and target languages. Table~\ref{tab:single_compatibility_main} reports detailed single-target results for the Slavic base, which covers all compatibility types in our design. Full Germanic and Romance per-target results are reported in Appendix~\ref{app:single_target_details}.

The results show that base--target compatibility affects expansion behavior, but not through a simple ranking of conditions. Serbian is the most compatible target in our design, with both the same language group and script as the Slavic base, and is comparatively easy to add. SSU-Wanda+TIES reaches 27.63\% target WER with only +4.17\% source degradation. However, same-group targets are not uniformly easy. Polish and Slovenian are both Slavic languages using Latin script in our setting, but they are harder to add than Serbian, and Slovenian causes much larger source degradation.

The comparison between Slovenian and Kyrgyz further shows that language-group match alone does not determine expansion difficulty. Slovenian is same-group but different-script, while Kyrgyz is different-group but same-script. Yet SSU-Wanda+TIES preserves source languages much better for Kyrgyz than for Slovenian, with +2.26\% versus +21.11\% source degradation, even though Kyrgyz remains difficult in target WER. This suggests that source retention and target learnability are related but not identical. A target can be hard to recognize while causing limited forgetting, or linguistically closer while producing a more disruptive update.

These mixed patterns suggest that group and script compatibility influence expansion, but their effect depends on how the target update interacts with source-important connector directions.

\subsection{Multi-Target Expansion}
\label{sec:multi_target}


Table~\ref{tab:multi_summary_main} reports final-stage results under the configured target order. The order is NO $\rightarrow$ IT $\rightarrow$ SR for Germanic and Romance, and IT $\rightarrow$ SR $\rightarrow$ PL $\rightarrow$ SL $\rightarrow$ KY for Slavic. Joint methods merge all target updates at the final stage, whereas recursive methods follow the configured order.

\begin{table}[!htbp]
\centering
\small
\setlength{\tabcolsep}{4pt}
\resizebox{\linewidth}{!}{%
\begin{tabular}{llrrr}
\toprule
\textbf{Base} & \textbf{Method} & \textbf{Source} & \textbf{Target} & \textbf{All} \\
\midrule
\multirow{6}{*}{GEM}
& Base & 22.57 & 102.70 & 52.62 \\
& TIES & 22.11 & 64.93 & \textbf{38.17} \\
& Hard SSU+TIES & 22.02 & 86.76 & 46.30 \\
& Incremental dynamic SSU & \textbf{21.64} & 84.26 & 45.12 \\
& Recursive rebased SSU & 37.13 & 45.39 & 40.23 \\
& Recursive FT then SSU & 28.97 & \textbf{57.34} & 39.61 \\
\midrule
\multirow{6}{*}{ROA}
& Base & 18.95 & 111.16 & 53.53 \\
& TIES & 19.58 & \textbf{67.22} & \textbf{37.45} \\
& Hard SSU+TIES & 18.10 & 98.67 & 48.32 \\
& Incremental dynamic SSU & \textbf{18.06} & 90.90 & 45.38 \\
& Recursive rebased SSU & 58.27 & 56.53 & 57.62 \\
& Recursive FT then SSU & 50.37 & 70.21 & 57.81 \\
\midrule
\multirow{6}{*}{SLA}
& Base & 36.93 & 110.15 & 73.54 \\
& TIES & \textbf{33.93} & 97.41 & 65.67 \\
& Hard SSU+TIES & 34.64 & 102.92 & 68.78 \\
& Incremental dynamic SSU & 33.91 & 98.63 & 66.27 \\
& Recursive rebased SSU & 103.43 & 73.61 & 88.52 \\
& Recursive FT then SSU & 38.45 & \textbf{81.67} & \textbf{60.06} \\
\bottomrule
\end{tabular}}
\caption{Final-stage WER (\%) for multi-target expansion. Lower is better.}
\label{tab:multi_summary_main}
\end{table}

\subsubsection{Joint Update Merging versus Recursive Adaptation}

Joint update merging provides a stable source-preserving solution, while recursive adaptation can learn target languages more aggressively but is more order-sensitive. TIES keeps source WER close to the base connector for all three base groups and substantially improves target WER, reducing it from 102.70\% to 64.93\% for Germanic, from 111.16\% to 67.22\% for Romance, and from 110.15\% to 97.41\% for Slavic. This shows that independently trained target updates can be merged jointly without severe source-language forgetting.

Source-aware variants of joint merging do not uniformly improve target performance in the multi-target setting. Hard SSU+TIES and Incremental dynamic SSU further protect source languages, but often retain less target-language improvement than TIES. This suggests that source protection can become too restrictive when multiple target updates are combined.

Recursive methods show a different trade-off. Recursive FT then SSU achieves the best target WER for Germanic and Slavic, and the best overall WER for Slavic under the configured order. However, its Romance result is much worse, indicating that recursive adaptation can produce stronger target learning but depends heavily on target order. Recursive rebased SSU is even less stable. Although it improves target WER in some settings, it causes large source degradation for Romance and Slavic, suggesting that recursively applying base-anchored target updates can accumulate source-language interference. Thus, unlike single-target expansion where SSU-Wanda+TIES is consistently strongest, multi-target expansion exposes a sharper trade-off between stable joint merging and order-sensitive recursive adaptation.

\subsubsection{Recursive Order Sensitivity}

\begin{table}[!htbp]
\centering
\small
\setlength{\tabcolsep}{4pt}

\resizebox{\linewidth}{!}{%
\begin{tabular}{llrrr}
\toprule
\textbf{Base} & \textbf{Target Order} & \textbf{Source} & \textbf{Target} & \textbf{All} \\
\midrule

\multirow{5}{*}{GEM}
& NO $\rightarrow$ IT $\rightarrow$ SR & 28.97 & 57.34 & 39.61 \\
& IT $\rightarrow$ NO $\rightarrow$ SR & 36.03 & 62.12 & 45.81 \\
& SR $\rightarrow$ IT $\rightarrow$ NO & \textbf{24.30} & 53.74 & 35.34 \\
& IT $\rightarrow$ SR $\rightarrow$ NO & 25.05 & 51.72 & \textbf{35.05} \\
& NO $\rightarrow$ SR $\rightarrow$ IT & 28.00 & \textbf{48.72} & 35.77 \\
\midrule

\multirow{5}{*}{ROA}
& NO $\rightarrow$ IT $\rightarrow$ SR & 50.37 & 70.21 & 57.81 \\
& IT $\rightarrow$ NO $\rightarrow$ SR & 23.57 & \textbf{45.59} & \textbf{31.82} \\
& SR $\rightarrow$ NO $\rightarrow$ IT & \textbf{21.07} & 59.16 & 35.35 \\
& NO $\rightarrow$ SR $\rightarrow$ IT & 23.83 & 54.37 & 35.28 \\
& IT $\rightarrow$ SR $\rightarrow$ NO & 23.00 & 52.35 & 34.00 \\
\midrule

\multirow{5}{*}{SLA}
& IT $\rightarrow$ SR $\rightarrow$ PL $\rightarrow$ SL $\rightarrow$ KY & \textbf{38.45} & 81.67 & \textbf{60.06} \\
& SR $\rightarrow$ PL $\rightarrow$ SL $\rightarrow$ KY $\rightarrow$ IT & 54.83 & \textbf{69.92} & 62.37 \\
& SR $\rightarrow$ KY $\rightarrow$ PL $\rightarrow$ SL $\rightarrow$ IT & 91.46 & 70.25 & 80.86 \\
& KY $\rightarrow$ PL $\rightarrow$ IT $\rightarrow$ SL $\rightarrow$ SR & 44.72 & 85.33 & 65.03 \\
& SR $\rightarrow$ SL $\rightarrow$ IT $\rightarrow$ PL $\rightarrow$ KY & 39.88 & 81.20 & 60.54 \\
\bottomrule
\end{tabular}}
\caption{Final-stage WER (\%) $\downarrow$ for Recursive FT then SSU under different orders.}
\label{tab:recursive_order_main}
\end{table}

Table~\ref{tab:recursive_order_main} reports Recursive FT then SSU under all evaluated target orders. The results show strong order sensitivity, so the configured-order result in Table~\ref{tab:multi_summary_main} should not be taken as the full capability of recursive adaptation.

For Romance, the configured order NO $\rightarrow$ IT $\rightarrow$ SR gives 57.81\% overall WER, while IT $\rightarrow$ NO $\rightarrow$ SR improves it to 31.82\%, outperforming TIES. Germanic shows a similar pattern, improving from 39.61\% under the configured order to 35.05\% with IT $\rightarrow$ SR $\rightarrow$ NO. For Slavic, Recursive FT then SSU already outperforms TIES under the configured order, but different orders still vary substantially, from 60.06\% to 80.86\% overall WER.

These results show that recursive adaptation is not simply worse than update merging. It is a higher-variance strategy that can achieve stronger target adaptation under favorable orders, but can also accumulate interference under poorly matched orders. This is consistent with the target-update conflict analysis in Appendix~\ref{app:target_update_conflict}, where target updates show low cosine similarity and substantial sign disagreement. Overall, update merging is more stable and source-preserving, while recursive adaptation can achieve better overall WER when the target sequence is favorable.

\section{Conclusion}
\label{sec:conclusion}

In this paper, we studied source-aware expansion for connector-based multilingual ASR, where new languages are added while preserving the original base connector. Our results show that target fine-tuning and standard TIES merging learn new languages but often overwrite source-language performance, while SSU-Wanda+TIES gives the strongest single-target adaptation--retention trade-off. For multi-target expansion, update merging is stable and source-preserving, whereas recursive adaptation can achieve stronger target performance under favorable orders but is highly order-sensitive. Overall, effective multilingual ASR expansion requires both source-aware preservation and careful coordination among target-language updates.

\section{Limitations}
\label{sec:limitations}

This work has two main limitations. First, our experiments focus on connector-based Speech LLMs, where the speech encoder and LLM are kept frozen and only the connector is adapted. This setting allows us to study parameter-selective adaptation and model merging in a controlled way, but the conclusions may not directly generalize to systems that update the speech encoder, the LLM, or the full ASR model during language expansion.
Second, our multi-target experiments cover a limited number of target languages and target orders. Although the results show clear order sensitivity in recursive expansion, we do not propose an automatic method for selecting the target order. A larger-scale study over more languages and principled order-selection strategies would be needed to fully characterize recursive multilingual expansion.

\section{Ethical Considerations}

Our experiments use publicly available speech data from Common Voice, without collecting new recordings or additional speaker metadata. We do not disclose any personally identifiable information about the speakers. Multilingual ASR expansion may broaden access to speech technologies, but improvements for newly added languages can come at the expense of previously supported languages, potentially reducing service quality for their speakers. ASR adaptation methods may affect languages and speaker groups differently, and improvements in average WER can obscure these disparities. To make differences across languages visible, we report per-language performance on both source and target languages.

\section{Acknowledgments}
This work has been funded by the European Union’s Horizon Europe research and innovation programme under the project ELOQUENCE (Grant Agreement No. 101135916) and the UKRI. Views and opinions expressed are, however, those of the author(s) only and do not necessarily reflect those of the European Union or the Research Executive Agency. Neither the European Union nor the granting authority can be held responsible for them.

\bibliography{ref}

\newpage

\newpage

\appendix

\section{Additional Experimental Setup}
\label{app:experimental_details}

\subsection{Data Allocation and Calibration}

Each multilingual base contains five source languages, with 5 hours of training speech per language, giving a total of 25 hours per base. Each target-language adaptation run uses 5 hours of speech. In single-target experiments, every run starts independently from the corresponding multilingual base connector.

The Germanic and Romance bases are each expanded to Norwegian, Italian, and Serbian. The Slavic base is expanded to Italian, Serbian, Polish, Slovenian, and Kyrgyz.

For SSU-Wanda, the calibration set contains 5 minutes of speech from the validation split of each source language, totaling 25 minutes per base. Calibration is therefore balanced by speech duration across source languages. It is used only for forward activation collection and mask construction. Source-language examples are not replayed or used for gradient-based adaptation during expansion. Random and magnitude-based masking do not require activation collection.

\subsection{Expansion Protocols}

We evaluate both single-target and multi-target expansion. In single-target expansion, each target language is added independently to a base connector, with base--target pairs covering language-group and script match/mismatch conditions (Table~\ref{tab:base_target_settings}). We compare direct target fine-tuning, TIES without source-aware protection, and SSU variants using random, magnitude-based, and Wanda-based masking.

In multi-target expansion, we study how multiple target-language updates can form a single expanded connector. For joint update merging, target-specific connectors are independently adapted from the same base connector. Each target update is computed relative to that shared base, and the updates are merged without further gradient-based training. Recursive adaptation instead processes targets sequentially, using the connector produced by the preceding stage as the starting point for the next target. Each stage uses only the current target language for gradient-based training.

The configured order is NO $\rightarrow$ IT $\rightarrow$ SR for the Germanic and Romance bases, and IT $\rightarrow$ SR $\rightarrow$ PL $\rightarrow$ SL $\rightarrow$ KY for the Slavic base. We additionally evaluate Recursive FT then SSU under the orders listed in Table~\ref{tab:recursive_order_main}. These constitute the evaluated subset of possible orders rather than an exhaustive enumeration.

\section{Additional Experimental Results}
\label{app:additional_results}

\subsection{Full Single-Target Expansion Results}
\label{app:single_target_details}

Tables~\ref{tab:ssu-single-ger}--\ref{tab:ssu-single-sla} report full per-language WER for independent single-target expansion. These results complement the macro-averaged summary in Table~\ref{tab:single_summary_main} and show the source-retention trends for each original source language.

\begin{table*}[p]
\centering
\footnotesize
\setlength{\tabcolsep}{4pt}
\begin{tabular}{@{}lrrrrrrrr@{}}
\toprule
& \multicolumn{5}{c}{Original source languages} & \multicolumn{1}{c}{Target} & \multicolumn{2}{c}{Source retention} \\
\cmidrule(lr){2-6}\cmidrule(lr){7-7}\cmidrule(l){8-9}
Method & DA & NL & EN & DE & SV & Target $\downarrow$ & Source $\downarrow$ & $\Delta$Source $\downarrow$ \\
\midrule
\multicolumn{9}{@{}l}{\textit{Added language: Norwegian}} \\
Base (zero-shot target) & 27.76 & 19.37 & 19.17 & 23.61 & 22.96 & 62.60 & 22.57 & +0.00 \\
Target FT & 44.52 & 19.99 & 17.48 & 24.26 & 62.70 & 33.12 & 33.79 & +11.21 \\
SSU-Wanda + FT (50\%) & 39.13 & 19.83 & 19.11 & 24.05 & 57.41 & 35.46 & 31.91 & +9.34 \\
TIES (no protection) & 45.08 & 19.94 & \textbf{17.72} & 24.26 & 61.88 & \textbf{33.08} & 33.78 & +11.20 \\
SSU-Rand + TIES & 32.47 & \textbf{18.91} & 18.06 & 23.38 & 35.73 & 36.59 & 25.71 & +3.14 \\
SSU-Mag + TIES & 35.48 & 19.39 & 18.17 & 23.69 & 45.16 & 36.23 & 28.38 & +5.80 \\
SSU-Wanda + TIES & \textbf{28.09} & 19.07 & 19.16 & \textbf{22.86} & \textbf{26.30} & 37.13 & \textbf{23.09} & \textbf{+0.52} \\
\addlinespace[4pt]
\multicolumn{9}{@{}l}{\textit{Added language: Italian}} \\
Base (zero-shot target) & 27.76 & 19.37 & 19.17 & 23.61 & 22.96 & 109.69 & 22.57 & +0.00 \\
Target FT & 37.70 & 86.12 & 18.93 & 70.51 & 71.32 & 18.66 & 56.92 & +34.34 \\
SSU-Wanda + FT (50\%) & 30.81 & 68.20 & 19.30 & 65.55 & 56.16 & 17.47 & 48.00 & +25.43 \\
TIES (no protection) & 35.54 & 82.52 & 18.80 & 68.39 & 69.35 & \textbf{18.44} & 54.92 & +32.34 \\
SSU-Rand + TIES & 27.12 & 19.93 & 18.52 & 23.20 & 24.90 & 18.76 & 22.74 & +0.16 \\
SSU-Mag + TIES & 28.20 & 23.00 & \textbf{18.44} & 25.69 & 29.21 & 18.71 & 24.91 & +2.34 \\
SSU-Wanda + TIES & \textbf{26.52} & \textbf{18.40} & 18.80 & \textbf{22.77} & \textbf{23.45} & 21.67 & \textbf{21.99} & \textbf{-0.59} \\
\addlinespace[4pt]
\multicolumn{9}{@{}l}{\textit{Added language: Serbian}} \\
Base (zero-shot target) & 27.76 & 19.37 & 19.17 & 23.61 & 22.96 & 135.82 & 22.57 & +0.00 \\
Target FT & 138.61 & 124.16 & 48.92 & 151.34 & 129.35 & 31.79 & 118.47 & +95.90 \\
SSU-Wanda + FT (50\%) & 147.62 & 132.01 & 44.59 & 153.74 & 134.36 & 32.48 & 122.46 & +99.89 \\
TIES (no protection) & 140.84 & 123.39 & 47.26 & 151.03 & 130.14 & \textbf{31.60} & 118.53 & +95.96 \\
SSU-Rand + TIES & 34.42 & 25.57 & 21.45 & 54.03 & 63.57 & 35.54 & 39.81 & +17.23 \\
SSU-Mag + TIES & 61.59 & 51.93 & 27.74 & 98.15 & 106.64 & 33.61 & 69.21 & +46.64 \\
SSU-Wanda + TIES & \textbf{28.91} & \textbf{18.65} & \textbf{17.68} & \textbf{25.57} & \textbf{26.00} & 48.68 & \textbf{23.36} & \textbf{+0.79} \\
\addlinespace[4pt]
\bottomrule
\end{tabular}
\caption{Germanic single-target expansion results. Values are WER (\%).}
\label{tab:ssu-single-ger}
\end{table*}

\begin{table*}[p]
\centering

\footnotesize
\setlength{\tabcolsep}{4pt}
\begin{tabular}{@{}lrrrrrrrr@{}}
\toprule
& \multicolumn{5}{c}{Original source languages} & \multicolumn{1}{c}{Target} & \multicolumn{2}{c}{Source retention} \\
\cmidrule(lr){2-6}\cmidrule(lr){7-7}\cmidrule(l){8-9}
Method & CA & FR & GL & PT & ES & Target $\downarrow$ & Source $\downarrow$ & $\Delta$Source $\downarrow$ \\
\midrule
\multicolumn{9}{@{}l}{\textit{Added language: Norwegian}} \\
Base (zero-shot target) & 26.11 & 20.20 & 22.69 & 13.28 & 12.48 & 124.03 & 18.95 & +0.00 \\
Target FT & 59.24 & 61.23 & 38.67 & 73.86 & 72.41 & 36.86 & 61.08 & +42.13 \\
SSU-Wanda + FT (50\%) & 31.41 & 32.66 & 26.57 & 33.10 & 24.28 & 41.04 & 29.61 & +10.66 \\
TIES (no protection) & 55.88 & 58.52 & 36.78 & 71.82 & 70.67 & \textbf{36.41} & 58.73 & +39.78 \\
SSU-Rand + TIES & 25.38 & 20.90 & 23.18 & 15.24 & 14.82 & 38.84 & 19.90 & +0.95 \\
SSU-Mag + TIES & 27.35 & 23.58 & 23.75 & 22.76 & 19.72 & 38.34 & 23.43 & +4.48 \\
SSU-Wanda + TIES & \textbf{24.41} & \textbf{19.78} & \textbf{22.42} & \textbf{12.73} & \textbf{12.18} & 46.94 & \textbf{18.30} & \textbf{-0.65} \\
\addlinespace[4pt]
\multicolumn{9}{@{}l}{\textit{Added language: Italian}} \\
Base (zero-shot target) & 26.11 & 20.20 & 22.69 & 13.28 & 12.48 & 83.57 & 18.95 & +0.00 \\
Target FT & 37.62 & 23.72 & 38.98 & 75.38 & 80.41 & 16.22 & 51.22 & +32.27 \\
SSU-Wanda + FT (50\%) & 29.10 & 21.11 & 31.38 & 61.98 & 69.41 & 16.09 & 42.60 & +23.65 \\
TIES (no protection) & 36.90 & 23.40 & 38.43 & 74.60 & 80.31 & 16.34 & 50.72 & +31.77 \\
SSU-Rand + TIES & 27.59 & 20.57 & 24.07 & 32.83 & 47.81 & 16.32 & 30.57 & +11.62 \\
SSU-Mag + TIES & 27.66 & 20.93 & 25.25 & 41.27 & 57.76 & \textbf{16.23} & 34.58 & +15.62 \\
SSU-Wanda + TIES & \textbf{24.82} & \textbf{20.08} & \textbf{22.37} & \textbf{14.56} & \textbf{16.59} & 17.06 & \textbf{19.69} & \textbf{+0.73} \\
\addlinespace[4pt]
\multicolumn{9}{@{}l}{\textit{Added language: Serbian}} \\
Base (zero-shot target) & 26.11 & 20.20 & 22.69 & 13.28 & 12.48 & 125.88 & 18.95 & +0.00 \\
Target FT & 108.78 & 115.21 & 108.97 & 108.39 & 114.58 & 31.10 & 111.19 & +92.23 \\
SSU-Wanda + FT (50\%) & 66.18 & 89.52 & 103.85 & 105.61 & 101.10 & 31.21 & 93.25 & +74.30 \\
TIES (no protection) & 106.13 & 114.98 & 111.46 & 109.02 & 114.92 & \textbf{30.57} & 111.30 & +92.35 \\
SSU-Rand + TIES & 35.34 & 42.33 & 48.10 & 75.70 & 52.99 & 33.35 & 50.89 & +31.94 \\
SSU-Mag + TIES & 46.21 & 59.95 & 62.15 & 86.74 & 73.00 & 33.35 & 65.61 & +46.66 \\
SSU-Wanda + TIES & \textbf{25.28} & \textbf{20.35} & \textbf{24.74} & \textbf{20.11} & \textbf{14.16} & 48.10 & \textbf{20.93} & \textbf{+1.97} \\
\addlinespace[4pt]
\bottomrule
\end{tabular}
\caption{Romance single-target expansion results. Values are WER (\%).}
\label{tab:ssu-single-rom}
\end{table*}
\clearpage

\begin{table*}[p]
\centering

\footnotesize
\setlength{\tabcolsep}{4pt}
\begin{tabular}{@{}lrrrrrrrr@{}}
\toprule
& \multicolumn{5}{c}{Original source languages} & \multicolumn{1}{c}{Target} & \multicolumn{2}{c}{Source retention} \\
\cmidrule(lr){2-6}\cmidrule(lr){7-7}\cmidrule(l){8-9}
Method & RU & UK & BE & BG & MK & Target $\downarrow$ & Source $\downarrow$ & $\Delta$Source $\downarrow$ \\
\midrule
\multicolumn{9}{@{}l}{\textit{Added language: Italian}} \\
Base (zero-shot target) & 27.79 & 52.88 & 37.52 & 31.48 & 34.97 & 116.08 & 36.93 & +0.00 \\
Target FT & 131.42 & 127.99 & 159.59 & 115.63 & 114.32 & 20.11 & 129.79 & +92.86 \\
SSU-Wanda + FT (50\%) & 108.31 & 113.37 & 133.94 & 111.24 & 109.35 & 23.36 & 115.24 & +78.31 \\
TIES (no protection) & 128.18 & 126.42 & 157.57 & 115.33 & 114.13 & \textbf{21.07} & 128.32 & +91.40 \\
SSU-Rand + TIES & 47.93 & 59.19 & 59.10 & 44.18 & 53.70 & 39.52 & 52.82 & +15.89 \\
SSU-Mag + TIES & 80.61 & 88.41 & 89.99 & 72.24 & 81.87 & 34.28 & 82.62 & +45.70 \\
SSU-Wanda + TIES & \textbf{21.77} & \textbf{39.48} & \textbf{40.76} & \textbf{32.37} & \textbf{37.38} & 70.57 & \textbf{34.35} & \textbf{-2.57} \\
\addlinespace[4pt]
\multicolumn{9}{@{}l}{\textit{Added language: Serbian}} \\
Base (zero-shot target) & 27.79 & 52.88 & 37.52 & 31.48 & 34.97 & 66.58 & 36.93 & +0.00 \\
Target FT & 46.53 & 66.46 & 53.93 & 63.81 & 54.87 & 24.51 & 57.12 & +20.19 \\
SSU-Wanda + FT (50\%) & 41.16 & 68.28 & 51.56 & 63.54 & 53.26 & 24.62 & 55.56 & +18.63 \\
TIES (no protection) & 45.51 & 65.73 & 52.85 & 63.46 & 54.74 & \textbf{24.62} & 56.46 & +19.53 \\
SSU-Rand + TIES & 35.40 & 45.33 & 40.17 & 54.56 & 45.84 & 26.03 & 44.26 & +7.33 \\
SSU-Mag + TIES & 36.96 & 47.55 & 41.23 & 57.41 & 47.45 & 25.91 & 46.12 & +9.19 \\
SSU-Wanda + TIES & \textbf{32.84} & \textbf{43.65} & \textbf{38.58} & \textbf{48.05} & \textbf{42.36} & 27.63 & \textbf{41.10} & \textbf{+4.17} \\
\addlinespace[4pt]
\multicolumn{9}{@{}l}{\textit{Added language: Polish}} \\
Base (zero-shot target) & 27.79 & 52.88 & 37.52 & 31.48 & 34.97 & 107.36 & 36.93 & +0.00 \\
Target FT & 107.16 & 106.96 & 109.19 & 106.82 & 105.14 & 26.39 & 107.05 & +70.13 \\
SSU-Wanda + FT (50\%) & 104.99 & 107.50 & 107.00 & 102.63 & 101.88 & 26.17 & 104.80 & +67.87 \\
TIES (no protection) & 107.27 & 106.91 & 109.82 & 107.07 & 105.72 & \textbf{26.28} & 107.36 & +70.43 \\
SSU-Rand + TIES & 80.16 & 104.34 & 83.35 & 50.13 & 66.77 & 28.09 & 76.95 & +40.02 \\
SSU-Mag + TIES & 98.06 & 106.02 & 98.56 & 82.44 & 90.75 & 27.42 & 95.16 & +58.24 \\
SSU-Wanda + TIES & \textbf{31.26} & \textbf{73.66} & \textbf{44.96} & \textbf{34.74} & \textbf{42.04} & 35.45 & \textbf{45.33} & \textbf{+8.41} \\
\addlinespace[4pt]
\multicolumn{9}{@{}l}{\textit{Added language: Slovenian}} \\
Base (zero-shot target) & 27.79 & 52.88 & 37.52 & 31.48 & 34.97 & 110.70 & 36.93 & +0.00 \\
Target FT & 104.34 & 106.17 & 105.25 & 105.95 & 106.29 & 38.47 & 105.60 & +68.67 \\
SSU-Wanda + FT (50\%) & 80.01 & 105.25 & 97.90 & 106.38 & 107.19 & 38.84 & 99.34 & +62.41 \\
TIES (no protection) & 103.98 & 105.29 & 105.04 & 105.85 & 106.66 & 40.07 & 105.37 & +68.44 \\
SSU-Rand + TIES & 49.48 & 87.00 & 66.82 & 100.91 & 105.76 & \textbf{39.99} & 81.99 & +45.07 \\
SSU-Mag + TIES & 70.86 & 98.35 & 83.38 & 105.27 & 106.80 & 40.03 & 92.93 & +56.00 \\
SSU-Wanda + TIES & \textbf{29.48} & \textbf{54.45} & \textbf{40.37} & \textbf{70.74} & \textbf{95.17} & 43.72 & \textbf{58.04} & \textbf{+21.11} \\
\addlinespace[4pt]
\multicolumn{9}{@{}l}{\textit{Added language: Kyrgyz}} \\
Base (zero-shot target) & 27.79 & 52.88 & 37.52 & 31.48 & 34.97 & 150.01 & 36.93 & +0.00 \\
Target FT & 29.25 & 77.96 & 79.61 & 42.80 & 57.93 & 72.39 & 57.51 & +20.58 \\
SSU-Wanda + FT (50\%) & 25.67 & 73.34 & 73.42 & 38.68 & 54.35 & 67.77 & 53.09 & +16.16 \\
TIES (no protection) & 29.19 & 77.97 & 78.43 & 42.14 & 56.71 & \textbf{73.09} & 56.89 & +19.96 \\
SSU-Rand + TIES & 22.84 & 69.78 & 43.75 & 31.47 & 38.66 & 74.84 & 41.30 & +4.37 \\
SSU-Mag + TIES & 23.11 & 71.78 & 47.83 & 32.84 & 41.66 & 75.09 & 43.45 & +6.52 \\
SSU-Wanda + TIES & \textbf{21.90} & \textbf{67.37} & \textbf{40.11} & \textbf{29.90} & \textbf{36.66} & 81.10 & \textbf{39.19} & \textbf{+2.26} \\
\addlinespace[4pt]
\bottomrule
\end{tabular}
\caption{Slavic single-target expansion results. Values are WER (\%).}
\label{tab:ssu-single-sla}
\end{table*}

\clearpage
\subsection{Full Multi-Target Expansion Results}
\label{app:multi_target_details}

Tables~\ref{tab:ssu-germanic-three-target-default}--\ref{tab:ssu-slavic-five-target-default} report stage-by-stage multi-target expansion results. These tables provide the per-language details behind the final-stage summary in Table~\ref{tab:multi_summary_main}.

\begin{table*}[p]
\centering
\scriptsize
\setlength{\tabcolsep}{2.5pt}
\resizebox{\textwidth}{!}{%
\begin{tabular}{@{}lllrrrrrrrrrrr@{}}
\toprule
Method & Stage & Accepted targets & DA $\downarrow$ & NL $\downarrow$ & EN $\downarrow$ & DE $\downarrow$ & SV $\downarrow$ & Source mean $\downarrow$ & NO $\downarrow$ & IT $\downarrow$ & SR $\downarrow$ & Target mean $\downarrow$ & All $\downarrow$ \\
\midrule
Base & 0 & Zero-shot & 27.76 & 19.37 & 19.17 & 23.61 & 22.96 & 22.57 & 62.60 & 109.69 & 135.82 & 102.70 & 52.62 \\
\midrule
\multicolumn{14}{@{}l}{\textit{Order: NO$\rightarrow$IT$\rightarrow$SR (configured)}} \\
\midrule
TIES (joint) & 1 & NO & 45.08 & 19.94 & 17.72 & 24.26 & 61.88 & 33.78 & 33.08 & -- & -- & 33.08 & 33.66 \\
TIES (joint) & 2 & NO$\rightarrow$IT & 27.88 & 18.84 & 17.46 & 23.48 & 24.44 & 22.42 & 36.99 & 40.15 & -- & 38.57 & 27.03 \\
TIES (joint) & 3 & NO$\rightarrow$IT$\rightarrow$SR & 27.91 & 18.52 & 16.83 & 22.25 & 25.03 & 22.11 & 36.41 & 40.65 & 117.72 & 64.93 & 38.17 \\
\addlinespace[2pt]
Hard SSU+TIES & 1 & NO & 28.09 & 19.07 & 19.16 & 22.86 & 26.30 & 23.09 & 37.13 & -- & -- & 37.13 & 25.43 \\
Hard SSU+TIES & 2 & NO$\rightarrow$IT & 26.55 & 19.03 & 18.72 & 22.82 & 23.13 & 22.05 & 53.69 & 86.24 & -- & 69.97 & 35.74 \\
Hard SSU+TIES & 3 & NO$\rightarrow$IT$\rightarrow$SR & 27.24 & 18.87 & 18.16 & 23.04 & 22.80 & 22.02 & 47.57 & 82.38 & 130.34 & 86.76 & 46.30 \\
\addlinespace[2pt]
Incremental dynamic SSU & 1 & NO & 28.15 & 19.23 & 19.34 & 22.48 & 26.65 & 23.17 & 36.95 & -- & -- & 36.95 & 25.47 \\
Incremental dynamic SSU & 2 & NO$\rightarrow$IT & 26.19 & 18.75 & 17.49 & 22.37 & 22.71 & 21.50 & 42.75 & 70.15 & -- & 56.45 & 31.49 \\
Incremental dynamic SSU & 3 & NO$\rightarrow$IT$\rightarrow$SR & 26.26 & 18.59 & 18.16 & 22.47 & 22.71 & 21.64 & 46.85 & 73.53 & 132.41 & 84.26 & 45.12 \\
\addlinespace[2pt]
Recursive rebased SSU & 1 & NO & 28.69 & 19.12 & 18.66 & 22.55 & 26.63 & 23.13 & 37.71 & -- & -- & 37.71 & 25.56 \\
Recursive rebased SSU & 2 & NO$\rightarrow$IT & 35.50 & 18.96 & 17.22 & 22.77 & 35.64 & 26.02 & 38.43 & 20.72 & -- & 29.58 & 27.03 \\
Recursive rebased SSU & 3 & NO$\rightarrow$IT$\rightarrow$SR & 43.00 & 26.48 & 17.69 & 39.85 & 58.62 & 37.13 & 43.83 & 31.53 & 60.79 & 45.39 & 40.23 \\
\addlinespace[2pt]
Recursive FT then SSU & 1 & NO & 28.18 & 19.11 & 18.82 & 22.89 & 26.62 & 23.13 & 37.35 & -- & -- & 37.35 & 25.50 \\
Recursive FT then SSU & 2 & NO$\rightarrow$IT & 31.35 & 18.59 & 17.90 & 22.43 & 27.27 & 23.51 & 39.96 & 20.89 & -- & 30.43 & 25.49 \\
Recursive FT then SSU & 3 & NO$\rightarrow$IT$\rightarrow$SR & 33.50 & 22.32 & 17.77 & 33.12 & 38.16 & 28.97 & 48.92 & 77.49 & 45.60 & 57.34 & 39.61 \\
\midrule
\multicolumn{14}{@{}l}{\textit{Order: IT$\rightarrow$NO$\rightarrow$SR}} \\
\midrule
TIES (joint) & 1 & IT & 35.54 & 82.52 & 18.80 & 68.39 & 69.35 & 54.92 & -- & 18.44 & -- & 18.44 & 48.84 \\
TIES (joint) & 2 & IT$\rightarrow$NO & 27.88 & 18.84 & 17.46 & 23.48 & 24.44 & 22.42 & 36.99 & 40.15 & -- & 38.57 & 27.03 \\
TIES (joint) & 3 & IT$\rightarrow$NO$\rightarrow$SR & 27.91 & 18.52 & 16.83 & 22.25 & 25.03 & 22.11 & 36.41 & 40.65 & 117.72 & 64.93 & 38.17 \\
\addlinespace[2pt]
Hard SSU+TIES & 1 & IT & 26.52 & 18.40 & 18.80 & 22.77 & 23.45 & 21.99 & -- & 21.67 & -- & 21.67 & 21.93 \\
Hard SSU+TIES & 2 & IT$\rightarrow$NO & 26.55 & 19.03 & 18.72 & 22.82 & 23.13 & 22.05 & 53.69 & 86.24 & -- & 69.97 & 35.74 \\
Hard SSU+TIES & 3 & IT$\rightarrow$NO$\rightarrow$SR & 27.24 & 18.87 & 18.16 & 23.04 & 22.80 & 22.02 & 47.57 & 82.38 & 130.34 & 86.76 & 46.30 \\
\addlinespace[2pt]
Incremental dynamic SSU & 1 & IT & 26.65 & 18.69 & 17.89 & 23.08 & 23.27 & 21.92 & -- & 21.18 & -- & 21.18 & 21.79 \\
Incremental dynamic SSU & 2 & IT$\rightarrow$NO & 26.08 & 18.77 & 18.25 & 22.65 & 22.78 & 21.70 & 43.11 & 62.40 & -- & 52.76 & 30.58 \\
Incremental dynamic SSU & 3 & IT$\rightarrow$NO$\rightarrow$SR & 26.50 & 18.78 & 18.33 & 22.54 & 22.66 & 21.76 & 43.97 & 65.97 & 132.25 & 80.73 & 43.87 \\
\addlinespace[2pt]
Recursive rebased SSU & 1 & IT & 26.69 & 18.63 & 18.47 & 23.17 & 23.14 & 22.02 & -- & 21.57 & -- & 21.57 & 21.95 \\
Recursive rebased SSU & 2 & IT$\rightarrow$NO & 35.54 & 19.14 & 17.02 & 23.03 & 35.31 & 26.01 & 35.19 & 20.29 & -- & 27.74 & 26.51 \\
Recursive rebased SSU & 3 & IT$\rightarrow$NO$\rightarrow$SR & 42.27 & 26.68 & 18.22 & 40.46 & 58.38 & 37.20 & 41.90 & 31.23 & 59.99 & 44.37 & 39.89 \\
\addlinespace[2pt]
Recursive FT then SSU & 1 & IT & 27.25 & 18.66 & 16.72 & 22.40 & 22.92 & 21.59 & -- & 20.53 & -- & 20.53 & 21.41 \\
Recursive FT then SSU & 2 & IT$\rightarrow$NO & 28.10 & 18.59 & 16.60 & 22.52 & 28.01 & 22.76 & 38.34 & 25.32 & -- & 31.83 & 25.35 \\
Recursive FT then SSU & 3 & IT$\rightarrow$NO$\rightarrow$SR & 34.89 & 27.72 & 19.49 & 50.52 & 47.51 & 36.03 & 66.88 & 81.67 & 37.81 & 62.12 & 45.81 \\
\midrule
\multicolumn{14}{@{}l}{\textit{Order: SR$\rightarrow$IT$\rightarrow$NO}} \\
\midrule
TIES (joint) & 1 & SR & 140.84 & 123.39 & 47.26 & 151.03 & 130.14 & 118.53 & -- & -- & 31.60 & 31.60 & 104.04 \\
TIES (joint) & 2 & SR$\rightarrow$IT & 26.27 & 18.43 & 17.90 & 22.32 & 23.76 & 21.73 & -- & 21.49 & 111.69 & 66.59 & 34.55 \\
TIES (joint) & 3 & SR$\rightarrow$IT$\rightarrow$NO & 27.91 & 18.52 & 16.83 & 22.25 & 25.03 & 22.11 & 36.41 & 40.65 & 117.72 & 64.93 & 38.17 \\
\addlinespace[2pt]
Hard SSU+TIES & 1 & SR & 28.91 & 18.65 & 17.68 & 25.57 & 26.00 & 23.36 & -- & -- & 48.68 & 48.68 & 27.58 \\
Hard SSU+TIES & 2 & SR$\rightarrow$IT & 26.31 & 18.58 & 17.82 & 22.48 & 23.06 & 21.65 & -- & 64.82 & 125.20 & 95.01 & 42.61 \\
Hard SSU+TIES & 3 & SR$\rightarrow$IT$\rightarrow$NO & 27.24 & 18.87 & 18.16 & 23.04 & 22.80 & 22.02 & 47.57 & 82.38 & 130.37 & 86.77 & 46.30 \\
\addlinespace[2pt]
Incremental dynamic SSU & 1 & SR & 28.65 & 18.61 & 17.99 & 25.73 & 26.25 & 23.44 & -- & -- & 49.25 & 49.25 & 27.75 \\
Incremental dynamic SSU & 2 & SR$\rightarrow$IT & 26.33 & 18.07 & 17.64 & 22.39 & 23.09 & 21.51 & -- & 56.40 & 113.91 & 85.15 & 39.69 \\
Incremental dynamic SSU & 3 & SR$\rightarrow$IT$\rightarrow$NO & 25.77 & 18.40 & 17.51 & 22.68 & 22.64 & 21.40 & 55.09 & 58.88 & 117.81 & 77.26 & 42.35 \\
\addlinespace[2pt]
Recursive rebased SSU & 1 & SR & 28.51 & 18.63 & 17.71 & 25.72 & 25.69 & 23.25 & -- & -- & 48.56 & 48.56 & 27.47 \\
Recursive rebased SSU & 2 & SR$\rightarrow$IT & 32.12 & 21.44 & 17.42 & 33.47 & 32.12 & 27.31 & -- & 43.42 & 49.35 & 46.39 & 32.76 \\
Recursive rebased SSU & 3 & SR$\rightarrow$IT$\rightarrow$NO & 42.79 & 26.41 & 17.77 & 39.87 & 58.34 & 37.04 & 41.22 & 32.54 & 58.93 & 44.23 & 39.73 \\
\addlinespace[2pt]
Recursive FT then SSU & 1 & SR & 31.27 & 20.39 & 19.37 & 36.68 & 41.01 & 29.74 & -- & -- & 35.68 & 35.68 & 30.73 \\
Recursive FT then SSU & 2 & SR$\rightarrow$IT & 29.59 & 18.65 & 16.75 & 23.04 & 26.60 & 22.93 & -- & 20.90 & 84.49 & 52.69 & 31.43 \\
Recursive FT then SSU & 3 & SR$\rightarrow$IT$\rightarrow$NO & 30.30 & 18.41 & 16.71 & 23.44 & 32.63 & 24.30 & 34.70 & 25.88 & 100.65 & 53.74 & 35.34 \\
\bottomrule
\end{tabular}%
}
\caption{Germanic multi-target expansion results under evaluated target orders. Values are WER (\%).}
\label{tab:ssu-germanic-three-target-default}
\end{table*}

\begin{table*}[p]
\centering
\scriptsize
\setlength{\tabcolsep}{2.5pt}
\resizebox{\textwidth}{!}{%
\begin{tabular}{@{}lllrrrrrrrrrrr@{}}
\toprule
Method & Stage & Accepted targets & CA $\downarrow$ & FR $\downarrow$ & GL $\downarrow$ & PT $\downarrow$ & ES $\downarrow$ & Source mean $\downarrow$ & NO $\downarrow$ & IT $\downarrow$ & SR $\downarrow$ & Target mean $\downarrow$ & All $\downarrow$ \\
\midrule
Base & 0 & Zero-shot & 26.11 & 20.20 & 22.69 & 13.28 & 12.48 & 18.95 & 124.03 & 83.57 & 125.88 & 111.16 & 53.53 \\
\midrule
\multicolumn{14}{@{}l}{\textit{Order: NO$\rightarrow$IT$\rightarrow$SR (configured)}} \\
\midrule
TIES (joint) & 1 & NO & 55.88 & 58.52 & 36.78 & 71.82 & 70.67 & 58.73 & 36.41 & -- & -- & 36.41 & 55.01 \\
TIES (joint) & 2 & NO$\rightarrow$IT & 24.56 & 19.58 & 22.43 & 13.91 & 17.38 & 19.57 & 71.83 & 18.17 & -- & 45.00 & 26.83 \\
TIES (joint) & 3 & NO$\rightarrow$IT$\rightarrow$SR & 24.21 & 19.55 & 22.53 & 14.59 & 17.04 & 19.58 & 68.00 & 19.13 & 114.52 & 67.22 & 37.45 \\
\addlinespace[2pt]
Hard SSU+TIES & 1 & NO & 24.41 & 19.78 & 22.42 & 12.73 & 12.18 & 18.30 & 46.94 & -- & -- & 46.94 & 23.08 \\
Hard SSU+TIES & 2 & NO$\rightarrow$IT & 24.38 & 19.69 & 22.44 & 12.73 & 12.08 & 18.26 & 123.54 & 53.79 & -- & 88.67 & 38.38 \\
Hard SSU+TIES & 3 & NO$\rightarrow$IT$\rightarrow$SR & 24.19 & 19.75 & 22.20 & 12.63 & 11.72 & 18.10 & 110.31 & 57.55 & 128.16 & 98.67 & 48.32 \\
\addlinespace[2pt]
Incremental dynamic SSU & 1 & NO & 24.41 & 19.73 & 22.41 & 13.31 & 12.72 & 18.52 & 45.32 & -- & -- & 45.32 & 22.98 \\
Incremental dynamic SSU & 2 & NO$\rightarrow$IT & 24.79 & 19.52 & 22.38 & 12.76 & 11.90 & 18.27 & 96.98 & 33.09 & -- & 65.04 & 31.63 \\
Incremental dynamic SSU & 3 & NO$\rightarrow$IT$\rightarrow$SR & 24.39 & 19.44 & 22.31 & 12.72 & 11.47 & 18.06 & 108.19 & 38.23 & 126.28 & 90.90 & 45.38 \\
\addlinespace[2pt]
Recursive rebased SSU & 1 & NO & 24.67 & 19.52 & 22.46 & 13.27 & 12.72 & 18.53 & 45.59 & -- & -- & 45.59 & 23.04 \\
Recursive rebased SSU & 2 & NO$\rightarrow$IT & 26.58 & 20.20 & 25.07 & 19.75 & 48.26 & 27.97 & 45.50 & 16.72 & -- & 31.11 & 28.87 \\
Recursive rebased SSU & 3 & NO$\rightarrow$IT$\rightarrow$SR & 37.62 & 26.78 & 68.59 & 79.70 & 78.66 & 58.27 & 64.45 & 52.50 & 52.65 & 56.53 & 57.62 \\
\addlinespace[2pt]
Recursive FT then SSU & 1 & NO & 24.72 & 19.89 & 22.65 & 12.81 & 12.71 & 18.56 & 46.31 & -- & -- & 46.31 & 23.18 \\
Recursive FT then SSU & 2 & NO$\rightarrow$IT & 24.84 & 19.38 & 23.03 & 14.47 & 20.14 & 20.37 & 57.92 & 16.22 & -- & 37.07 & 25.14 \\
Recursive FT then SSU & 3 & NO$\rightarrow$IT$\rightarrow$SR & 33.96 & 28.44 & 46.95 & 83.08 & 59.41 & 50.37 & 106.93 & 68.18 & 35.51 & 70.21 & 57.81 \\
\midrule
\multicolumn{14}{@{}l}{\textit{Order: IT$\rightarrow$NO$\rightarrow$SR}} \\
\midrule
TIES (joint) & 1 & IT & 36.90 & 23.40 & 38.43 & 74.60 & 80.31 & 50.72 & -- & 16.34 & -- & 16.34 & 44.99 \\
TIES (joint) & 2 & IT$\rightarrow$NO & 24.56 & 19.58 & 22.43 & 13.91 & 17.38 & 19.57 & 71.83 & 18.17 & -- & 45.00 & 26.83 \\
TIES (joint) & 3 & IT$\rightarrow$NO$\rightarrow$SR & 24.21 & 19.55 & 22.53 & 14.59 & 17.04 & 19.58 & 68.00 & 19.13 & 114.52 & 67.22 & 37.45 \\
\addlinespace[2pt]
Hard SSU+TIES & 1 & IT & 24.82 & 20.08 & 22.37 & 14.56 & 16.59 & 19.69 & -- & 17.06 & -- & 17.06 & 19.25 \\
Hard SSU+TIES & 2 & IT$\rightarrow$NO & 24.38 & 19.69 & 22.44 & 12.73 & 12.08 & 18.26 & 123.54 & 53.79 & -- & 88.67 & 38.38 \\
Hard SSU+TIES & 3 & IT$\rightarrow$NO$\rightarrow$SR & 24.19 & 19.75 & 22.20 & 12.63 & 11.72 & 18.10 & 110.31 & 57.55 & 128.16 & 98.67 & 48.32 \\
\addlinespace[2pt]
Incremental dynamic SSU & 1 & IT & 25.22 & 20.34 & 22.82 & 14.55 & 16.75 & 19.93 & -- & 16.88 & -- & 16.88 & 19.43 \\
Incremental dynamic SSU & 2 & IT$\rightarrow$NO & 24.21 & 19.78 & 22.54 & 12.80 & 11.92 & 18.25 & 103.51 & 27.66 & -- & 65.59 & 31.78 \\
Incremental dynamic SSU & 3 & IT$\rightarrow$NO$\rightarrow$SR & 24.18 & 19.79 & 22.45 & 12.76 & 11.80 & 18.19 & 103.11 & 32.51 & 124.36 & 86.66 & 43.87 \\
\addlinespace[2pt]
Recursive rebased SSU & 1 & IT & 25.52 & 20.07 & 22.38 & 14.42 & 16.58 & 19.79 & -- & 16.98 & -- & 16.98 & 19.32 \\
Recursive rebased SSU & 2 & IT$\rightarrow$NO & 27.06 & 20.51 & 25.29 & 20.02 & 49.77 & 28.53 & 47.75 & 16.20 & -- & 31.98 & 29.51 \\
Recursive rebased SSU & 3 & IT$\rightarrow$NO$\rightarrow$SR & 38.53 & 26.97 & 69.64 & 79.00 & 78.89 & 58.61 & 59.77 & 52.87 & 52.98 & 55.21 & 57.33 \\
\addlinespace[2pt]
Recursive FT then SSU & 1 & IT & 24.80 & 19.35 & 23.32 & 13.61 & 14.49 & 19.11 & -- & 16.86 & -- & 16.86 & 18.74 \\
Recursive FT then SSU & 2 & IT$\rightarrow$NO & 24.77 & 19.45 & 22.94 & 13.28 & 17.12 & 19.51 & 50.68 & 17.68 & -- & 34.18 & 23.70 \\
Recursive FT then SSU & 3 & IT$\rightarrow$NO$\rightarrow$SR & 25.54 & 21.22 & 28.81 & 18.45 & 23.81 & 23.57 & 60.40 & 23.99 & 52.37 & 45.59 & 31.82 \\
\midrule
\multicolumn{14}{@{}l}{\textit{Order: SR$\rightarrow$NO$\rightarrow$IT}} \\
\midrule
TIES (joint) & 1 & SR & 106.13 & 114.98 & 111.46 & 109.02 & 114.92 & 111.30 & -- & -- & 30.57 & 30.57 & 97.84 \\
TIES (joint) & 2 & SR$\rightarrow$NO & 24.74 & 20.69 & 24.78 & 24.51 & 18.60 & 22.66 & 41.13 & -- & 90.87 & 66.00 & 35.05 \\
TIES (joint) & 3 & SR$\rightarrow$NO$\rightarrow$IT & 24.19 & 19.53 & 22.56 & 14.59 & 17.01 & 19.58 & 64.04 & 19.19 & 114.93 & 66.05 & 37.01 \\
\addlinespace[2pt]
Hard SSU+TIES & 1 & SR & 25.28 & 20.35 & 24.74 & 20.11 & 14.16 & 20.93 & -- & -- & 48.10 & 48.10 & 25.46 \\
Hard SSU+TIES & 2 & SR$\rightarrow$NO & 24.06 & 19.22 & 22.08 & 12.62 & 11.67 & 17.93 & 99.91 & -- & 127.42 & 113.66 & 45.28 \\
Hard SSU+TIES & 3 & SR$\rightarrow$NO$\rightarrow$IT & 24.15 & 19.76 & 22.19 & 12.65 & 11.72 & 18.09 & 109.81 & 57.60 & 128.39 & 98.60 & 48.28 \\
\addlinespace[2pt]
Incremental dynamic SSU & 1 & SR & 25.05 & 20.03 & 24.80 & 20.55 & 14.23 & 20.93 & -- & -- & 47.55 & 47.55 & 25.37 \\
Incremental dynamic SSU & 2 & SR$\rightarrow$NO & 23.68 & 19.70 & 22.74 & 13.00 & 12.19 & 18.26 & 82.00 & -- & 110.25 & 96.12 & 40.51 \\
Incremental dynamic SSU & 3 & SR$\rightarrow$NO$\rightarrow$IT & 23.62 & 19.40 & 22.84 & 12.88 & 12.38 & 18.22 & 86.00 & 65.77 & 116.46 & 89.41 & 44.92 \\
\addlinespace[2pt]
Recursive rebased SSU & 1 & SR & 25.10 & 19.98 & 24.71 & 20.29 & 13.97 & 20.81 & -- & -- & 47.81 & 47.81 & 25.31 \\
Recursive rebased SSU & 2 & SR$\rightarrow$NO & 33.60 & 29.30 & 45.92 & 69.17 & 49.07 & 45.41 & 75.43 & -- & 41.72 & 58.58 & 49.17 \\
Recursive rebased SSU & 3 & SR$\rightarrow$NO$\rightarrow$IT & 38.54 & 27.46 & 70.50 & 80.22 & 80.78 & 59.50 & 63.46 & 54.52 & 49.70 & 55.89 & 58.15 \\
\addlinespace[2pt]
Recursive FT then SSU & 1 & SR & 24.81 & 21.55 & 26.36 & 25.69 & 15.80 & 22.84 & -- & -- & 43.17 & 43.17 & 26.23 \\
Recursive FT then SSU & 2 & SR$\rightarrow$NO & 24.10 & 20.45 & 25.01 & 17.56 & 14.64 & 20.35 & 41.09 & -- & 69.51 & 55.30 & 30.34 \\
Recursive FT then SSU & 3 & SR$\rightarrow$NO$\rightarrow$IT & 25.05 & 20.41 & 25.41 & 15.20 & 19.27 & 21.07 & 63.50 & 16.60 & 97.38 & 59.16 & 35.35 \\
\bottomrule
\end{tabular}%
}
\caption{Romance multi-target expansion results under evaluated target orders. Values are WER (\%).}
\label{tab:ssu-romance-three-target-orders}
\end{table*}

\begin{table*}[p]
\centering

\scriptsize
\setlength{\tabcolsep}{2pt}
\resizebox{\textwidth}{!}{%
\begin{tabular}{@{}lllrrrrrrrrrrrrr@{}}
\toprule
Method & Stage & Accepted targets & RU $\downarrow$ & UK $\downarrow$ & BE $\downarrow$ & BG $\downarrow$ & MK $\downarrow$ & Source mean $\downarrow$ & IT $\downarrow$ & SR $\downarrow$ & PL $\downarrow$ & SL $\downarrow$ & KY $\downarrow$ & Target mean $\downarrow$ & All $\downarrow$ \\
\midrule
Base & 0 & Zero-shot & 27.79 & 52.88 & 37.52 & 31.48 & 34.97 & 36.93 & 116.08 & 66.58 & 107.36 & 110.70 & 150.01 & 110.15 & 73.54 \\
\midrule
\multicolumn{16}{@{}l}{\textit{Order: IT$\rightarrow$SR$\rightarrow$PL$\rightarrow$SL$\rightarrow$KY (configured)}} \\
\midrule
TIES (joint) & 1 & IT & 128.18 & 126.42 & 157.57 & 115.33 & 114.13 & 128.32 & 21.07 & -- & -- & -- & -- & 21.07 & 110.45 \\
TIES (joint) & 2 & IT$\rightarrow$SR & 23.17 & 34.90 & 39.05 & 34.31 & 35.79 & 33.44 & 85.05 & 43.92 & -- & -- & -- & 64.49 & 42.31 \\
TIES (joint) & 3 & IT$\rightarrow$SR$\rightarrow$PL & 34.41 & 68.00 & 47.44 & 35.09 & 41.76 & 45.34 & 71.05 & 69.67 & 42.62 & -- & -- & 61.11 & 51.26 \\
TIES (joint) & 4 & IT$\rightarrow$SR$\rightarrow$PL$\rightarrow$SL & 24.78 & 40.90 & 38.96 & 33.32 & 38.27 & 35.25 & 99.18 & 60.48 & 90.83 & 95.93 & -- & 86.60 & 58.07 \\
TIES (joint) & 5 & IT$\rightarrow$SR$\rightarrow$PL$\rightarrow$SL$\rightarrow$KY & 23.22 & 41.70 & 37.33 & 31.01 & 36.40 & 33.93 & 100.49 & 56.15 & 99.63 & 100.40 & 130.37 & 97.41 & 65.67 \\
\addlinespace[2pt]
Hard SSU+TIES & 1 & IT & 21.77 & 39.48 & 40.76 & 32.37 & 37.38 & 34.35 & 70.57 & -- & -- & -- & -- & 70.57 & 40.39 \\
Hard SSU+TIES & 2 & IT$\rightarrow$SR & 24.49 & 41.86 & 37.12 & 31.48 & 34.70 & 33.93 & 108.69 & 51.72 & -- & -- & -- & 80.21 & 47.15 \\
Hard SSU+TIES & 3 & IT$\rightarrow$SR$\rightarrow$PL & 23.09 & 41.03 & 37.21 & 31.49 & 34.23 & 33.41 & 108.26 & 55.74 & 105.01 & -- & -- & 89.67 & 54.51 \\
Hard SSU+TIES & 4 & IT$\rightarrow$SR$\rightarrow$PL$\rightarrow$SL & 25.48 & 45.28 & 36.73 & 31.38 & 34.42 & 34.66 & 111.25 & 58.99 & 106.05 & 107.27 & -- & 95.89 & 61.87 \\
Hard SSU+TIES & 5 & IT$\rightarrow$SR$\rightarrow$PL$\rightarrow$SL$\rightarrow$KY & 24.65 & 46.94 & 36.77 & 30.90 & 33.95 & 34.64 & 110.54 & 59.66 & 105.35 & 108.82 & 130.24 & 102.92 & 68.78 \\
\addlinespace[2pt]
Incremental dynamic SSU & 1 & IT & 21.72 & 39.42 & 40.16 & 32.41 & 37.71 & 34.28 & 70.94 & -- & -- & -- & -- & 70.94 & 40.39 \\
Incremental dynamic SSU & 2 & IT$\rightarrow$SR & 24.90 & 39.21 & 37.31 & 33.97 & 35.37 & 34.15 & 107.04 & 40.93 & -- & -- & -- & 73.99 & 45.53 \\
Incremental dynamic SSU & 3 & IT$\rightarrow$SR$\rightarrow$PL & 24.85 & 39.26 & 37.32 & 33.36 & 34.78 & 33.91 & 106.99 & 44.17 & 104.57 & -- & -- & 85.24 & 53.16 \\
Incremental dynamic SSU & 4 & IT$\rightarrow$SR$\rightarrow$PL$\rightarrow$SL & 24.85 & 39.26 & 37.32 & 33.36 & 34.78 & 33.91 & 106.99 & 44.17 & 104.57 & 108.12 & -- & 90.96 & 59.27 \\
Incremental dynamic SSU & 5 & IT$\rightarrow$SR$\rightarrow$PL$\rightarrow$SL$\rightarrow$KY & 24.85 & 39.26 & 37.32 & 33.36 & 34.78 & 33.91 & 106.99 & 44.17 & 104.57 & 108.12 & 129.30 & 98.63 & 66.27 \\
\addlinespace[2pt]
Recursive rebased SSU & 1 & IT & 21.72 & 39.25 & 40.33 & 32.41 & 38.15 & 34.37 & 70.05 & -- & -- & -- & -- & 70.05 & 40.32 \\
Recursive rebased SSU & 2 & IT$\rightarrow$SR & 25.71 & 35.06 & 42.31 & 45.25 & 43.39 & 38.34 & 69.07 & 34.78 & -- & -- & -- & 51.93 & 42.22 \\
Recursive rebased SSU & 3 & IT$\rightarrow$SR$\rightarrow$PL & 96.49 & 105.05 & 106.12 & 99.59 & 100.21 & 101.49 & 41.33 & 101.54 & 32.34 & -- & -- & 58.40 & 85.33 \\
Recursive rebased SSU & 4 & IT$\rightarrow$SR$\rightarrow$PL$\rightarrow$SL & 104.34 & 104.03 & 107.08 & 103.87 & 105.00 & 104.86 & 54.26 & 102.69 & 39.56 & 46.23 & -- & 60.69 & 85.23 \\
Recursive rebased SSU & 5 & IT$\rightarrow$SR$\rightarrow$PL$\rightarrow$SL$\rightarrow$KY & 100.35 & 103.26 & 105.79 & 103.32 & 104.44 & 103.43 & 49.69 & 102.69 & 51.00 & 46.47 & 118.19 & 73.61 & 88.52 \\
\addlinespace[2pt]
Recursive FT then SSU & 1 & IT & 21.20 & 40.66 & 40.84 & 32.17 & 38.72 & 34.72 & 69.59 & -- & -- & -- & -- & 69.59 & 40.53 \\
Recursive FT then SSU & 2 & IT$\rightarrow$SR & 32.92 & 39.23 & 43.06 & 50.56 & 47.51 & 42.66 & 104.13 & 25.59 & -- & -- & -- & 64.86 & 49.00 \\
Recursive FT then SSU & 3 & IT$\rightarrow$SR$\rightarrow$PL & 81.45 & 99.21 & 73.48 & 49.38 & 53.14 & 71.33 & 82.84 & 51.58 & 29.19 & -- & -- & 54.53 & 65.03 \\
Recursive FT then SSU & 4 & IT$\rightarrow$SR$\rightarrow$PL$\rightarrow$SL & 96.66 & 96.77 & 96.15 & 100.96 & 105.33 & 99.17 & 80.10 & 102.49 & 47.25 & 39.82 & -- & 67.41 & 85.06 \\
Recursive FT then SSU & 5 & IT$\rightarrow$SR$\rightarrow$PL$\rightarrow$SL$\rightarrow$KY & 22.22 & 47.88 & 45.83 & 34.63 & 41.71 & 38.45 & 86.83 & 44.14 & 99.92 & 97.37 & 80.07 & 81.67 & 60.06 \\
\midrule
\multicolumn{16}{@{}l}{\textit{Order: SR$\rightarrow$PL$\rightarrow$SL$\rightarrow$KY$\rightarrow$IT}} \\
\midrule
TIES (joint) & 1 & SR & 45.51 & 65.73 & 52.85 & 63.46 & 54.74 & 56.46 & -- & 24.62 & -- & -- & -- & 24.62 & 51.15 \\
TIES (joint) & 2 & SR$\rightarrow$PL & 48.07 & 83.15 & 49.36 & 39.54 & 42.44 & 52.51 & -- & 60.89 & 33.89 & -- & -- & 47.39 & 51.05 \\
TIES (joint) & 3 & SR$\rightarrow$PL$\rightarrow$SL & 50.69 & 86.72 & 57.45 & 64.85 & 76.11 & 67.16 & -- & 96.02 & 41.44 & 55.00 & -- & 64.15 & 66.03 \\
TIES (joint) & 4 & SR$\rightarrow$PL$\rightarrow$SL$\rightarrow$KY & 24.93 & 45.85 & 36.67 & 30.78 & 34.35 & 34.52 & -- & 51.43 & 104.50 & 105.06 & 123.07 & 96.02 & 61.85 \\
TIES (joint) & 5 & SR$\rightarrow$PL$\rightarrow$SL$\rightarrow$KY$\rightarrow$IT & 22.99 & 41.69 & 37.46 & 30.95 & 36.51 & 33.92 & 101.45 & 56.23 & 99.59 & 99.55 & 128.48 & 97.06 & 65.49 \\
\addlinespace[2pt]
Hard SSU+TIES & 1 & SR & 32.84 & 43.65 & 38.58 & 48.05 & 42.36 & 41.10 & -- & 27.63 & -- & -- & -- & 27.63 & 38.85 \\
Hard SSU+TIES & 2 & SR$\rightarrow$PL & 24.82 & 42.45 & 36.97 & 32.83 & 35.39 & 34.49 & -- & 50.02 & 105.27 & -- & -- & 77.65 & 46.82 \\
Hard SSU+TIES & 3 & SR$\rightarrow$PL$\rightarrow$SL & 25.98 & 44.13 & 36.80 & 32.87 & 35.93 & 35.14 & -- & 50.70 & 104.74 & 106.28 & -- & 87.24 & 54.68 \\
Hard SSU+TIES & 4 & SR$\rightarrow$PL$\rightarrow$SL$\rightarrow$KY & 25.70 & 49.63 & 37.03 & 31.15 & 34.40 & 35.58 & -- & 58.63 & 106.60 & 107.22 & 135.16 & 101.90 & 65.06 \\
Hard SSU+TIES & 5 & SR$\rightarrow$PL$\rightarrow$SL$\rightarrow$KY$\rightarrow$IT & 24.66 & 46.91 & 36.51 & 30.64 & 33.89 & 34.52 & 110.19 & 59.61 & 106.07 & 107.85 & 131.02 & 102.95 & 68.74 \\
\addlinespace[2pt]
Incremental dynamic SSU & 1 & SR & 32.65 & 43.97 & 38.64 & 48.25 & 42.65 & 41.23 & -- & 27.90 & -- & -- & -- & 27.90 & 39.01 \\
Incremental dynamic SSU & 2 & SR$\rightarrow$PL & 25.73 & 39.80 & 37.18 & 35.56 & 37.60 & 35.17 & -- & 38.29 & 104.98 & -- & -- & 71.64 & 45.59 \\
Incremental dynamic SSU & 3 & SR$\rightarrow$PL$\rightarrow$SL & 25.67 & 40.96 & 37.09 & 35.05 & 37.16 & 35.19 & -- & 40.46 & 104.49 & 107.62 & -- & 84.19 & 53.56 \\
Incremental dynamic SSU & 4 & SR$\rightarrow$PL$\rightarrow$SL$\rightarrow$KY & 25.67 & 40.96 & 37.09 & 35.05 & 37.16 & 35.19 & -- & 40.46 & 104.49 & 107.62 & 136.58 & 97.29 & 62.79 \\
Incremental dynamic SSU & 5 & SR$\rightarrow$PL$\rightarrow$SL$\rightarrow$KY$\rightarrow$IT & 25.67 & 40.96 & 37.09 & 35.05 & 37.16 & 35.19 & 110.26 & 40.46 & 104.49 & 107.62 & 136.58 & 99.88 & 67.53 \\
\addlinespace[2pt]
Recursive rebased SSU & 1 & SR & 32.83 & 43.54 & 38.48 & 48.18 & 42.51 & 41.11 & -- & 27.92 & -- & -- & -- & 27.92 & 38.91 \\
Recursive rebased SSU & 2 & SR$\rightarrow$PL & 59.75 & 92.70 & 58.64 & 56.97 & 52.18 & 64.05 & -- & 47.21 & 37.64 & -- & -- & 42.43 & 57.87 \\
Recursive rebased SSU & 3 & SR$\rightarrow$PL$\rightarrow$SL & 106.48 & 105.58 & 107.29 & 104.99 & 105.64 & 106.00 & -- & 102.86 & 57.74 & 43.51 & -- & 68.04 & 91.76 \\
Recursive rebased SSU & 4 & SR$\rightarrow$PL$\rightarrow$SL$\rightarrow$KY & 92.82 & 103.61 & 104.91 & 103.14 & 104.70 & 101.84 & -- & 102.71 & 68.79 & 44.71 & 122.93 & 84.78 & 94.26 \\
Recursive rebased SSU & 5 & SR$\rightarrow$PL$\rightarrow$SL$\rightarrow$KY$\rightarrow$IT & 100.37 & 103.10 & 105.16 & 103.31 & 104.34 & 103.26 & 51.81 & 102.66 & 51.28 & 47.13 & 120.83 & 74.74 & 89.00 \\
\addlinespace[2pt]
Recursive FT then SSU & 1 & SR & 47.22 & 49.35 & 41.32 & 56.31 & 48.08 & 48.46 & -- & 26.40 & -- & -- & -- & 26.40 & 44.78 \\
Recursive FT then SSU & 2 & SR$\rightarrow$PL & 63.93 & 90.68 & 56.17 & 54.52 & 52.06 & 63.47 & -- & 39.12 & 34.37 & -- & -- & 36.74 & 55.84 \\
Recursive FT then SSU & 3 & SR$\rightarrow$PL$\rightarrow$SL & 98.46 & 100.15 & 96.28 & 101.57 & 104.40 & 100.17 & -- & 100.54 & 48.50 & 39.90 & -- & 62.98 & 86.23 \\
Recursive FT then SSU & 4 & SR$\rightarrow$PL$\rightarrow$SL$\rightarrow$KY & 23.83 & 49.46 & 43.10 & 35.60 & 42.19 & 38.84 & -- & 43.54 & 99.34 & 97.20 & 84.63 & 81.17 & 57.65 \\
Recursive FT then SSU & 5 & SR$\rightarrow$PL$\rightarrow$SL$\rightarrow$KY$\rightarrow$IT & 32.06 & 59.30 & 61.97 & 48.74 & 72.07 & 54.83 & 46.52 & 83.36 & 61.34 & 64.05 & 94.32 & 69.92 & 62.37 \\
\midrule
\multicolumn{16}{@{}l}{\textit{Order: SR$\rightarrow$KY$\rightarrow$PL$\rightarrow$SL$\rightarrow$IT}} \\
\midrule
TIES (joint) & 1 & SR & 45.51 & 65.73 & 52.85 & 63.46 & 54.74 & 56.46 & -- & 24.62 & -- & -- & -- & 24.62 & 51.15 \\
TIES (joint) & 2 & SR$\rightarrow$KY & 25.95 & 53.47 & 37.06 & 31.40 & 33.62 & 36.30 & -- & 38.72 & -- & -- & 104.53 & 71.62 & 46.39 \\
TIES (joint) & 3 & SR$\rightarrow$KY$\rightarrow$PL & 22.75 & 43.95 & 36.90 & 30.54 & 34.56 & 33.74 & -- & 46.26 & 101.85 & -- & 116.81 & 88.31 & 54.20 \\
TIES (joint) & 4 & SR$\rightarrow$KY$\rightarrow$PL$\rightarrow$SL & 24.94 & 45.85 & 36.50 & 30.79 & 34.35 & 34.49 & -- & 51.48 & 104.51 & 105.06 & 123.11 & 96.04 & 61.84 \\
TIES (joint) & 5 & SR$\rightarrow$KY$\rightarrow$PL$\rightarrow$SL$\rightarrow$IT & 23.25 & 41.71 & 37.34 & 31.00 & 36.43 & 33.94 & 100.71 & 56.14 & 99.59 & 100.36 & 129.91 & 97.34 & 65.64 \\
\addlinespace[2pt]
Hard SSU+TIES & 1 & SR & 32.84 & 43.65 & 38.58 & 48.05 & 42.36 & 41.10 & -- & 27.63 & -- & -- & -- & 27.63 & 38.85 \\
Hard SSU+TIES & 2 & SR$\rightarrow$KY & 25.42 & 53.61 & 37.32 & 30.53 & 33.87 & 36.15 & -- & 51.01 & -- & -- & 124.74 & 87.88 & 50.93 \\
Hard SSU+TIES & 3 & SR$\rightarrow$KY$\rightarrow$PL & 24.58 & 48.10 & 36.77 & 30.36 & 34.24 & 34.81 & -- & 56.08 & 105.77 & -- & 129.70 & 97.18 & 58.20 \\
Hard SSU+TIES & 4 & SR$\rightarrow$KY$\rightarrow$PL$\rightarrow$SL & 25.70 & 49.63 & 37.02 & 31.14 & 34.40 & 35.58 & -- & 58.59 & 106.59 & 107.19 & 135.14 & 101.88 & 65.05 \\
Hard SSU+TIES & 5 & SR$\rightarrow$KY$\rightarrow$PL$\rightarrow$SL$\rightarrow$IT & 24.65 & 46.93 & 36.77 & 30.90 & 33.95 & 34.64 & 110.39 & 59.64 & 105.36 & 108.80 & 130.21 & 102.88 & 68.76 \\
\addlinespace[2pt]
Incremental dynamic SSU & 1 & SR & 32.65 & 43.97 & 38.64 & 48.25 & 42.65 & 41.23 & -- & 27.90 & -- & -- & -- & 27.90 & 39.01 \\
Incremental dynamic SSU & 2 & SR$\rightarrow$KY & 25.37 & 51.90 & 36.85 & 30.84 & 33.61 & 35.71 & -- & 41.77 & -- & -- & 113.98 & 77.88 & 47.76 \\
Incremental dynamic SSU & 3 & SR$\rightarrow$KY$\rightarrow$PL & 25.03 & 50.98 & 36.83 & 30.58 & 33.95 & 35.47 & -- & 43.84 & 105.03 & -- & 119.28 & 89.38 & 55.69 \\
Incremental dynamic SSU & 4 & SR$\rightarrow$KY$\rightarrow$PL$\rightarrow$SL & 25.03 & 50.98 & 36.83 & 30.58 & 33.95 & 35.47 & -- & 43.84 & 105.03 & 106.99 & 119.28 & 93.79 & 61.39 \\
Incremental dynamic SSU & 5 & SR$\rightarrow$KY$\rightarrow$PL$\rightarrow$SL$\rightarrow$IT & 25.03 & 50.98 & 36.83 & 30.58 & 33.95 & 35.47 & 111.01 & 43.84 & 105.03 & 106.99 & 119.28 & 97.23 & 66.35 \\
\addlinespace[2pt]
Recursive rebased SSU & 1 & SR & 32.83 & 43.54 & 38.48 & 48.18 & 42.51 & 41.11 & -- & 27.92 & -- & -- & -- & 27.92 & 38.91 \\
Recursive rebased SSU & 2 & SR$\rightarrow$KY & 25.28 & 54.85 & 39.00 & 35.27 & 36.31 & 38.14 & -- & 29.66 & -- & -- & 88.58 & 59.12 & 44.14 \\
Recursive rebased SSU & 3 & SR$\rightarrow$KY$\rightarrow$PL & 21.83 & 45.97 & 40.98 & 39.13 & 41.85 & 37.95 & -- & 33.21 & 86.68 & -- & 91.99 & 70.63 & 50.20 \\
Recursive rebased SSU & 4 & SR$\rightarrow$KY$\rightarrow$PL$\rightarrow$SL & 89.81 & 103.48 & 104.35 & 103.06 & 104.61 & 101.06 & -- & 102.73 & 70.24 & 44.86 & 119.75 & 84.39 & 93.65 \\
Recursive rebased SSU & 5 & SR$\rightarrow$KY$\rightarrow$PL$\rightarrow$SL$\rightarrow$IT & 100.15 & 103.48 & 107.57 & 103.42 & 104.24 & 103.77 & 51.91 & 102.25 & 51.95 & 47.08 & 119.75 & 74.59 & 89.18 \\
\addlinespace[2pt]
Recursive FT then SSU & 1 & SR & 43.27 & 53.60 & 40.85 & 51.83 & 45.10 & 46.93 & -- & 26.53 & -- & -- & -- & 26.53 & 43.53 \\
Recursive FT then SSU & 2 & SR$\rightarrow$KY & 26.58 & 62.20 & 42.26 & 33.91 & 36.26 & 40.24 & -- & 33.84 & -- & -- & 81.15 & 57.50 & 45.17 \\
Recursive FT then SSU & 3 & SR$\rightarrow$KY$\rightarrow$PL & 23.17 & 50.99 & 42.30 & 33.40 & 40.54 & 38.08 & -- & 36.42 & 93.22 & -- & 86.93 & 72.19 & 50.87 \\
Recursive FT then SSU & 4 & SR$\rightarrow$KY$\rightarrow$PL$\rightarrow$SL & 68.37 & 86.41 & 77.26 & 82.64 & 97.89 & 82.51 & -- & 98.30 & 56.94 & 43.03 & 121.35 & 79.91 & 81.36 \\
Recursive FT then SSU & 5 & SR$\rightarrow$KY$\rightarrow$PL$\rightarrow$SL$\rightarrow$IT & 80.73 & 88.71 & 95.41 & 90.37 & 102.08 & 91.46 & 40.58 & 101.25 & 40.34 & 43.93 & 125.15 & 70.25 & 80.86 \\
\bottomrule
\end{tabular}%
}
\caption{Slavic multi-target expansion results under evaluated target orders. Values are WER (\%).}
\label{tab:ssu-slavic-five-target-default}
\end{table*}

\clearpage
\subsection{Recursive Multi-Target Order}
\label{app:recursive_order}

Table~\ref{tab:ssu-all-families-mergeability-orders-five-hour} reports the full stage-by-stage trajectories for the recursive orders summarized in Table~\ref{tab:recursive_order_main}. Target mean is computed over accepted targets at each stage.

\begin{table*}[p]
\centering
\scriptsize
\setlength{\tabcolsep}{2.5pt}
\renewcommand{\arraystretch}{1.0}
\resizebox{\textwidth}{!}{%
\begin{tabular}{@{}lllrrrrrrrrrrr@{}}
\toprule
\multicolumn{14}{@{}l}{\textbf{Romance}} \\
\midrule
Method & Stage & Accepted targets & CA $\downarrow$ & FR $\downarrow$ & GL $\downarrow$ & PT $\downarrow$ & ES $\downarrow$ & Source mean $\downarrow$ & NO $\downarrow$ & IT $\downarrow$ & SR $\downarrow$ & Target mean $\downarrow$ & All $\downarrow$ \\
\midrule
Base & 0 & Zero-shot & 26.11 & 20.20 & 22.69 & 13.28 & 12.48 & 18.95 & 124.03 & 83.57 & 125.88 & 111.16 & 53.53 \\
\midrule
\multicolumn{14}{@{}l}{\textit{Easy-to-hard: NO$\rightarrow$SR$\rightarrow$IT}} \\
\midrule
Recursive FT then SSU & 1 & NO & 24.21 & 19.50 & 22.59 & 13.05 & 12.29 & 18.33 & 46.80 & -- & -- & 46.80 & 23.07 \\
Recursive FT then SSU & 2 & NO$\rightarrow$SR & 42.22 & 48.04 & 52.57 & 88.80 & 65.94 & 59.52 & 108.24 & -- & 32.83 & 70.53 & 62.66 \\
Recursive FT then SSU & 3 & NO$\rightarrow$SR$\rightarrow$IT & 26.58 & 21.12 & 27.72 & 19.33 & 24.37 & 23.83 & 72.73 & 17.00 & 73.38 & 54.37 & 35.28 \\
\midrule
\multicolumn{14}{@{}l}{\textit{Hard-to-easy: IT$\rightarrow$SR$\rightarrow$NO}} \\
\midrule
Recursive FT then SSU & 1 & IT & 25.09 & 19.54 & 23.02 & 13.64 & 14.54 & 19.17 & -- & 16.88 & -- & 16.88 & 18.78 \\
Recursive FT then SSU & 2 & IT$\rightarrow$SR & 25.51 & 21.15 & 28.24 & 21.02 & 21.28 & 23.44 & -- & 21.13 & 53.30 & 37.22 & 27.38 \\
Recursive FT then SSU & 3 & IT$\rightarrow$SR$\rightarrow$NO & 26.93 & 21.45 & 27.84 & 16.90 & 21.87 & 23.00 & 44.73 & 25.47 & 86.84 & 52.35 & 34.00 \\
\bottomrule
\end{tabular}%
}
\par\smallskip
\resizebox{\textwidth}{!}{%
\begin{tabular}{@{}lllrrrrrrrrrrr@{}}
\toprule
\multicolumn{14}{@{}l}{\textbf{Germanic}} \\
\midrule
Method & Stage & Accepted targets & DA $\downarrow$ & NL $\downarrow$ & EN $\downarrow$ & DE $\downarrow$ & SV $\downarrow$ & Source mean $\downarrow$ & NO $\downarrow$ & IT $\downarrow$ & SR $\downarrow$ & Target mean $\downarrow$ & All $\downarrow$ \\
\midrule
Base & 0 & Zero-shot & 27.76 & 19.37 & 19.17 & 23.61 & 22.96 & 22.57 & 62.60 & 109.69 & 135.82 & 102.70 & 52.62 \\
\midrule
\multicolumn{14}{@{}l}{\textit{Easy-to-hard: IT$\rightarrow$SR$\rightarrow$NO}} \\
\midrule
Recursive FT then SSU & 1 & IT & 26.93 & 18.24 & 16.75 & 22.91 & 22.89 & 21.54 & -- & 20.28 & -- & 20.28 & 21.33 \\
Recursive FT then SSU & 2 & IT$\rightarrow$SR & 46.09 & 29.61 & 21.17 & 55.76 & 53.78 & 41.28 & -- & 98.45 & 35.96 & 67.21 & 48.69 \\
Recursive FT then SSU & 3 & IT$\rightarrow$SR$\rightarrow$NO & 29.68 & 18.94 & 18.80 & 23.81 & 34.03 & 25.05 & 34.83 & 43.63 & 76.70 & 51.72 & 35.05 \\
\midrule
\multicolumn{14}{@{}l}{\textit{Hard-to-easy: NO$\rightarrow$SR$\rightarrow$IT}} \\
\midrule
Recursive FT then SSU & 1 & NO & 27.88 & 19.10 & 17.28 & 22.65 & 26.00 & 22.58 & 40.86 & -- & -- & 40.86 & 25.63 \\
Recursive FT then SSU & 2 & NO$\rightarrow$SR & 35.08 & 29.03 & 19.20 & 48.54 & 49.10 & 36.19 & 57.25 & -- & 36.75 & 47.00 & 39.28 \\
Recursive FT then SSU & 3 & NO$\rightarrow$SR$\rightarrow$IT & 37.40 & 20.15 & 18.35 & 24.90 & 39.17 & 28.00 & 38.16 & 20.48 & 87.51 & 48.72 & 35.77 \\
\bottomrule
\end{tabular}%
}
\par\smallskip
\resizebox{\textwidth}{!}{%
\begin{tabular}{@{}lllrrrrrrrrrrrrr@{}}
\toprule
\multicolumn{16}{@{}l}{\textbf{Slavic}} \\
\midrule
Method & Stage & Accepted targets & RU $\downarrow$ & UK $\downarrow$ & BE $\downarrow$ & BG $\downarrow$ & MK $\downarrow$ & Source mean $\downarrow$ & IT $\downarrow$ & SR $\downarrow$ & PL $\downarrow$ & SL $\downarrow$ & KY $\downarrow$ & Target mean $\downarrow$ & All $\downarrow$ \\
\midrule
Base & 0 & Zero-shot & 27.79 & 52.88 & 37.52 & 31.48 & 34.97 & 36.93 & 116.08 & 66.58 & 107.36 & 110.70 & 150.01 & 110.15 & 73.54 \\
\midrule
\multicolumn{16}{@{}l}{\textit{Easy-to-hard: KY$\rightarrow$PL$\rightarrow$IT$\rightarrow$SL$\rightarrow$SR}} \\
\midrule
Recursive FT then SSU & 1 & KY & 21.84 & 65.68 & 41.93 & 30.80 & 37.42 & 39.53 & -- & -- & -- & -- & 74.97 & 74.97 & 45.44 \\
Recursive FT then SSU & 2 & KY$\rightarrow$PL & 19.68 & 54.32 & 43.10 & 30.71 & 39.29 & 37.42 & -- & -- & 93.31 & -- & 86.82 & 90.07 & 52.46 \\
Recursive FT then SSU & 3 & KY$\rightarrow$PL$\rightarrow$IT & 32.87 & 68.13 & 67.02 & 40.10 & 60.47 & 53.72 & 36.94 & -- & 49.37 & -- & 116.98 & 67.76 & 58.98 \\
Recursive FT then SSU & 4 & KY$\rightarrow$PL$\rightarrow$IT$\rightarrow$SL & 66.34 & 85.65 & 89.81 & 78.42 & 102.17 & 84.48 & 53.17 & -- & 35.76 & 40.15 & 123.59 & 63.17 & 75.01 \\
Recursive FT then SSU & 5 & KY$\rightarrow$PL$\rightarrow$IT$\rightarrow$SL$\rightarrow$SR & 28.57 & 43.43 & 48.20 & 51.88 & 51.53 & 44.72 & 88.41 & 29.84 & 91.13 & 99.56 & 117.72 & 85.33 & 65.03 \\
\midrule
\multicolumn{16}{@{}l}{\textit{Hard-to-easy: SR$\rightarrow$SL$\rightarrow$IT$\rightarrow$PL$\rightarrow$KY}} \\
\midrule
Recursive FT then SSU & 1 & SR & 35.25 & 46.35 & 41.10 & 50.36 & 47.03 & 44.02 & -- & 25.52 & -- & -- & -- & 25.52 & 40.93 \\
Recursive FT then SSU & 2 & SR$\rightarrow$SL & 53.48 & 75.17 & 59.03 & 94.10 & 97.06 & 75.77 & -- & 88.45 & -- & 46.09 & -- & 67.27 & 73.34 \\
Recursive FT then SSU & 3 & SR$\rightarrow$SL$\rightarrow$IT & 53.80 & 80.22 & 82.95 & 96.74 & 104.90 & 83.72 & 41.06 & 99.93 & -- & 40.89 & -- & 60.63 & 75.06 \\
Recursive FT then SSU & 4 & SR$\rightarrow$SL$\rightarrow$IT$\rightarrow$PL & 88.77 & 104.89 & 102.80 & 100.58 & 104.07 & 100.22 & 62.36 & 103.36 & 28.04 & 56.91 & -- & 62.67 & 83.53 \\
Recursive FT then SSU & 5 & SR$\rightarrow$SL$\rightarrow$IT$\rightarrow$PL$\rightarrow$KY & 21.04 & 48.75 & 46.61 & 35.25 & 47.75 & 39.88 & 82.16 & 55.15 & 96.78 & 90.72 & 81.19 & 81.20 & 60.54 \\
\bottomrule
\end{tabular}%
}
\caption{Stage-by-stage Recursive FT then SSU results under additional target orders. Values are WER (\%).}
\label{tab:ssu-all-families-mergeability-orders-five-hour}
\end{table*}

\clearpage
\subsection{Target-Update Conflict}
\label{app:target_update_conflict}

Table~\ref{tab:target_conflict_appendix} reports pairwise target-update conflict after TIES trimming. Low cosine similarity and high sign disagreement indicate that target updates often point in conflicting directions.

\begin{table*}[htbp]
\centering
\footnotesize
\setlength{\tabcolsep}{4pt}

\resizebox{\textwidth}{!}{%
\begin{tabular}{lllccc}
\toprule
\textbf{Base} & \textbf{Target A} & \textbf{Target B} & \textbf{Cosine} & \textbf{Sign Disagreement} & \textbf{Support Jaccard} \\
\midrule
Germanic & Norwegian & Italian & 0.053 & 0.473 & 0.667 \\
Germanic & Norwegian & Serbian & 0.037 & 0.481 & 0.667 \\
Germanic & Italian & Serbian & 0.047 & 0.475 & 0.667 \\
\midrule
Romance & Norwegian & Italian & 0.064 & 0.466 & 0.668 \\
Romance & Norwegian & Serbian & 0.062 & 0.468 & 0.667 \\
Romance & Italian & Serbian & 0.048 & 0.475 & 0.667 \\
\midrule
Slavic & Italian & Serbian & 0.034 & 0.483 & 0.667 \\
Slavic & Italian & Polish & 0.096 & 0.447 & 0.668 \\
Slavic & Italian & Slovenian & 0.063 & 0.466 & 0.668 \\
Slavic & Italian & Kyrgyz & 0.032 & 0.485 & 0.667 \\
Slavic & Serbian & Polish & 0.045 & 0.476 & 0.667 \\
Slavic & Serbian & Slovenian & 0.066 & 0.465 & 0.668 \\
Slavic & Serbian & Kyrgyz & 0.034 & 0.484 & 0.667 \\
Slavic & Polish & Slovenian & 0.082 & 0.455 & 0.668 \\
Slavic & Polish & Kyrgyz & 0.019 & 0.492 & 0.667 \\
Slavic & Slovenian & Kyrgyz & 0.036 & 0.482 & 0.667 \\
\bottomrule
\end{tabular}}
\caption{Pairwise target-update conflict after TIES trimming.}
\label{tab:target_conflict_appendix}
\end{table*}

\end{document}